\documentclass[final,5p,times,twocolumn,authoryear]{elsarticle}
\usepackage{amssymb}
\usepackage{amsmath}
\usepackage[table]{xcolor}
\usepackage{tikz} 
\usepackage{tabularx}
\usepackage{booktabs}
\usepackage{multirow}
\usepackage{makecell}
\usepackage{subfig}
\usepackage{mathrsfs}
\usepackage{graphicx}
\usepackage[hyphens]{url}   % allow breaks at hyphens and slashes
\usepackage{hyperref}
\hypersetup{breaklinks=true}
\usepackage{colortbl}
\definecolor{lightgray}{gray}{0.6}

\newcolumntype{Y}{>{\raggedright\arraybackslash}X}
\newcolumntype{Z}{>{\centering\arraybackslash}X}
\newcommand{\notes}[1]{}  % TODO remove this and all notes before submission ~~~~~~~~~~~~~~~~~~~~

\journal{Medical Image Analysis}

\begin{document}

\begin{frontmatter}



% \title{Point-Cloud Registration for Liver Navigation using a Biomechanical Attention Network} %% Article title

% \title{Aligning the unseen: Non-rigid Volume-to-Surface Registration for Liver Navigation using a Biomechanical-Aware Point Transformer Network} %% Article title

\title{PIVOTS: Aligning unseen Structures using Preoperative to Intraoperative Volume-To-Surface Registration for Liver Navigation}


\author[nct,dkfz,ukd,hzdr]{Peng Liu\corref{cor1}\fnref{equalcontrib}} %% Author name
\author[nct,dkfz,ukd,hzdr]{Bianca Güttner\fnref{equalcontrib}} 
\author[nct,dkfz,ukd,hzdr]{Yutong Su}
\author[nct,dkfz,ukd,hzdr]{Chenyang Li}
\author[nct,dkfz,ukd,hzdr]{Jinjing Xu}
\author[sdu]{Mingyang Liu}
\author[sdu]{Zhe Min}
\author[nki_sd]{Andrey Zhylka}
\author[nki_sd]{Jasper Smit}
\author[nki_sd]{Karin Olthof}
\author[nki_sd]{Matteo Fusaglia} 
\author[tio,ukd,dkfz,hrdr]{Rudi Apolle}
\author[tio,ukd,dkfz,hrdr]{Matthias Miederer}
\author[ukd]{Laura Frohneberger}
\author[ukd]{Carina Riediger}
\author[ukd]{Jürgen Weitz}
\author[ukd_vis,purdue]{Fiona Kolbinger}
\author[nct,dkfz,ukd,hzdr,ceti]{Stefanie Speidel}
\author[nct,dkfz,ukd,hzdr]{Micha Pfeiffer}

\affiliation[nct]{organization={Translational Surgical Oncology, National Center for Tumor Diseases},%Department and Organization
            addressline={Fetscherstrasse 74/PF 64}, 
            city={Dresden},
            postcode={01307}, 
            state={Saxony},
            country={Germany}}

\affiliation[dkfz]{organization={German Cancer Research Center (DKFZ)},
            addressline={Im Neuenheimer Feld 280}, 
            city={Heidelberg},
            postcode={69120}, 
            state={Baden-Württemberg},
            country={Germany}}

\affiliation[ukd]{organization={Faculty of Medicine and University Hospital Carl Gustav Carus},
            city={Dresden},
            postcode={01307},             
            state={Saxony},
            country={Germany}}

\affiliation[hzdr]{organization={Helmholtz-Zentrum Dresden-Rossendorf (HZDR)},
            city={Dresden},
            state={Saxony},
            country={Germany}}

\affiliation[ukd_vis]{organization={Department of Visceral, Thoracic and Vascular Surgery, University Hospital and Faculty of Medicine Carl Gustav Carus, TUD Dresden University of Technology},
            city={Dresden},
            postcode={01307},             
            state={Saxony},
            country={Germany}}

\affiliation[ceti]{organization={Centre for Tactile Internet with Human-in-the-Loop, TU Dresden},
            city={Dresden},
            state={Saxony},
            country={Germany}}

\affiliation[sdu]{organization={School of Control Science and Engineering, Shandong University},
            city={Jinan},
            state={Shandong},
            country={China}}

\affiliation[nki_sd]{organization={Surgical Department, The Netherlands Cancer Institute},
            city={Amsterdam},
            country={Netherlands}}

\affiliation[tio]{organization={Translational Imaging in Oncology, National Center for Tumor Diseases},
            addressline={Fetscherstrasse 74/PF 64}, 
            city={Dresden},
            postcode={01307}, 
            state={Saxony},
            country={Germany}}
            
% \affiliation[nki_mp]{organization={Medical Physics, The Netherlands Cancer Institute},
%             % addressline={}, 
%             city={Amsterdam},
%             % postcode={},             
%             % state={},
%             country={Netherlands}}

\affiliation[purdue]{organization={Weldon School of Biomedical Engineering, Purdue University},
            city={West Lafayette, IN},
            country={USA}}

% Weldon School of Biomedical Engineering, Purdue University, West Lafayette, IN, USA

\cortext[cor1]{Corresponding author.
\textit{Email:} peng.liu@nct-dresden.de}

\fntext[equalcontrib]{These authors contributed equally to this work.}

%% Abstract
\begin{abstract}
% why intraoperative registration - special challenges - this leads to specific requirements - how we address them very roughly - I we show competing performance to baseline methods: our registration works, II we show effectiveness of our design to address the specific requirements with ablations, III we demonstrate qualitatively on real intraoperative data. - this leaves avenues A, B, C open for further research

Non-rigid registration is essential for Augmented Reality guided laparoscopic liver surgery by fusing preoperative information, such as tumor location and vascular structures, into the limited intraoperative view, thereby enhancing surgical navigation. A prerequisite is the accurate prediction of intraoperative liver deformation which remains highly challenging due to factors such as large deformation caused by  pneumoperitoneum, respiration and tool interaction as well as noisy intraoperative data, and limited field of view due to occlusion and constrained camera movement. To address these challenges, we introduce \textbf{PIVOTS}, a \textbf{P}reoperative to \textbf{I}ntraoperative \textbf{VO}lume-\textbf{T}o-\textbf{S}urface registration neural network that directly takes point clouds as input for deformation prediction. The geometric feature extraction encoder allows multi-resolution feature extraction, and the decoder, comprising novel deformation aware cross attention modules, enables pre- and intraoperative information interaction and accurate multi-level displacement prediction.
% Our model consists of an encoder for point downsampling and geometric feature propagation and a decoder employing deformation aware point transformer for information interaction and upsampling. 
We train the neural network on synthetic data simulated from a biomechanical simulation pipeline and validate its performance  on both synthetic and real datasets. Results demonstrate superior registration performance of our method compared to baseline methods, exhibiting strong robustness against high amounts of noise, large deformation, and various levels of intraoperative visibility. We publish the training and test sets as evaluation benchmarks and call for a fair comparison of liver registration methods with volume-to-surface data. 
% We also believe the proposed network could be applied in deformation prediction of other non-rigid soft tissues. 
Code and datasets are available here \url{https://github.com/pengliu-nct/PIVOTS}.

\end{abstract}

% % \frame{
% % \includegraphics[trim=0cm 17.5cm 1.1cm 0cm,clip,width=\textwidth]{pics/graphical_abstract_500x200_9_libre.pdf}
% \includegraphics[trim=0cm 0.3cm 1.2cm 0cm,clip,width=\textwidth]{pics/graphical_abstract_500x200_11_libre.pdf}
% % }
% % \includegraphics[trim=0cm 17.3cm 1.7cm 0cm,clip,width=\textwidth]{pics/arch_6.pdf}

% 85 Characters maximum
% \begin{highlights}
% \item Research highlight 1: Trained solely on synthetic simulation data and achieves zero-shot transfer to real patient data with no additional training
% \item Research highlight 2: Robust volume-to-surface registration method even when presented with challenges such as large deformation, partial visibility and substantial noise
% \item Research highlight 3: Designed a soft-tissue simulation pipeline to generate custom training and evaluation data for registration methods
% \item Research highlight 4: Curated a high-quality real human liver breathing motion dataset as a benchmark for evaluating 3D registration methods
% \end{highlights}

%% Keywords

\end{frontmatter}

% Document ToDos ~~~~~~~~~~~~~~~~~~~~~~~~~~~~~~~~~~~~~~~~~~~~~~~~~~~~~~~~~~~~~~~~~~~~~~~~~~~~~~~~~~~~~~~~~~~~~~~~~~~~~~~~~~~~~
% 
% - at the end: go through comments below and adapt document to requirements
% - check citation style \(\rightarrow\) Elsevier's Harvard referencing style (in-text citation sorting consistent, reference list with journal abbreviation, edition, pages and DOI)
% 
% 
% 
% 
% 
% 
% 


% main text
%%

\section{Introduction}
\label{sec:introduction}
%% Labels are used to cross-reference an item using \ref command.

\begin{figure}
\end{figure}
\begin{figure*}[!ht]
    \centering
    % \includegraphics[trim=0cm 13cm 0cm 0cm,clip,width=\textwidth]{pics/teaser_.pdf}
    % \frame{
    \includegraphics[trim=0cm 14.5cm 2.01cm 0cm,clip,width=\textwidth]{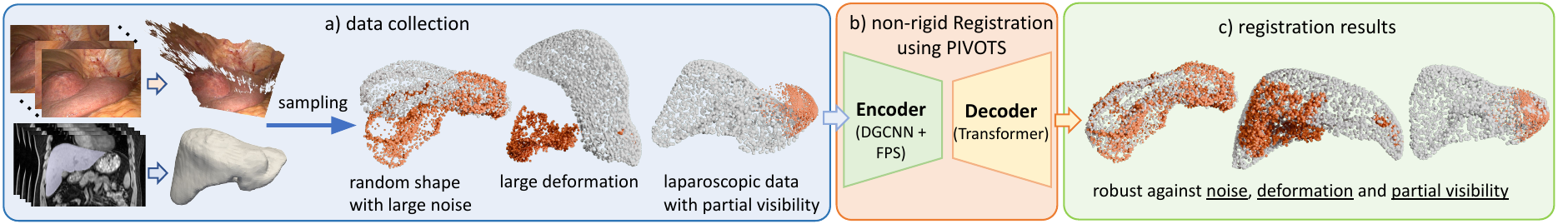}
    % }
    \caption{Registration with our method. First, the intraoperative surface (orange) is reconstructed from laparoscopic video  and the preoperative volume (gray) is segmented from MRI, and the surface and volume are subsampled as point clouds, depicted in a). These point clouds are then fed into PIVOTS for deformation prediction in b). Finally, the registration results in c) show the robustness of PIVOTS against substantial noise, large deformation and partial visibility.}
    \label{fig:teaser}
\end{figure*}

Laparoscopic liver surgery (LLS) is a minimally invasive approach for hepatic resection that utilizes small abdominal incisions, specialized instruments, and high-resolution camera systems. During LLS, surgeons strategically insert trocars into the abdominal cavity, through which they introduce surgical instruments and a laparoscope for real-time imaging, allowing them to navigate and resect the targeted liver tissue. Compared to open surgery, LLS offers decreased blood loss, postoperative pain and infectious complications, shorter hospital stays, and faster recovery, while matching the oncological results with respect to margin accuracy and recurrence rates 
% \cite{novitsky_net_2004},\cite{abu_hilal_southampton_2018}.
\citep{novitsky_net_2004, abu_hilal_southampton_2018, amodu_oncologic_2022, nguyen2011comparative, ratti2018laparoscopic}. 

Despite these advantages, such minimally invasive procedures also introduce multiple challenges. Intraoperative views only show the tissue surface that presents scarce distinct texture features, are often restricted by occlusions from adjacent tissues or instruments and confined operating spaces, and additionally degraded by camera artifacts and light reflections. 
At the same time, preoperative MRI or CT scans provide a comprehensive overview of the abdominal anatomy, detailing critical internal anatomical structures, such as vessels and tumors. Each patient presents a different setting due to morphological variations~\citep{singh2019study, srimani2020liver, santhosh2024study} 
% where parts may be missing from previous operations like in one of the training examples of the P2ILF challenge \citep{ali2025objective}, 
or due to large tumors distorting the natural shape of the organ. Besides instrument interaction, during surgery, the liver undergoes significant deformation due to respiratory motion, cardiac pulsation, pneumoperitoneum pressure, the round ligament stabilizing the left lobe and contact with other organs. The tissue properties of a patient's liver, e.g. stiffness, are unknown prior to surgery while they hold a big influence on the intraoperative deformation and show great variability across the population and across measurement techniques and conditions \citep{mattei_sample_2016, marchesseau_nonlinear_2017, murad_gutierrez_liver_2018, lemine_mechanical_2024}: Healthy liver tissue is soft and pliable with a Young's modulus below $2.5$\,kPa, becoming much stiffer with comorbidities like liver cirrhosis or fatty liver disease, with ranges up to $16$\,kPa in non-invasive elastography. These unknowns and mismatches between the data collection modalities impede the direct transfer of preoperative information to intraoperative settings. 

However, accurate information transfer is critical for patient safety, since liver surgery involves complex and variable vascular anatomy (hepatic arteries, portal and hepatic veins) and bile duct anatomy, the identification of which can be challenging, particularly in the limited view setting of LLS. Injuries to these structures can lead to serious complications like bleeding, reduced blood supply to the remaining liver, and, potentially, liver failure. 
Therefore, image-guided laparoscopic navigation systems with augmented reality (AR) are designed to facilitate orientation. Such a navigation system must satisfy several key requirements: It must \textit{i)} ensure an accurate alignment of pre- and intraoperative structures, reliably identifying the visible surface part, \textit{ii)} infer a precise, detailed and sufficiently smooth online deformation of critical structures (such as vessels and small metastases) from the visible parts of the organ,% which is detailed and smooth enough to capture the fine critical structures,  
\textit{iii)} demonstrate a high robustness against intraoperative conditions like noise and require \textit{iv)} limited manual intervention while \textit{v)} running on available hardware at very high speeds. To meet these demands, a navigation system must rely on a registration algorithm that is both precise and computationally efficient. 

3D-3D registration methods utilize the preoperative liver volume or surface and align it to an intraoperative 3D surface. This intraoperative target surface can be constructed using various methods, such as \textit{Structure from Motion} (SfM) or \textit{Simultaneous Localization and Mapping} (SLAM) methods \citep{Docea2021, Docea2022} or by tracking the tool tips \citep{Heiselman2024}. Such reconstructed surfaces are typically small with little overlap with the preoperative full liver volume, and are subject to different types of noise compared to the segmented preoperative 3D liver surfaces. Noise properties differ between reconstruction pipelines based on the choice of endoscope, calibration method, segmentation method, error accumulation due to heating, tracking and mapping method, amongst others. 
Rigid registration algorithms then aim to find the optimal rigid transformation between the pre- and intraoperative surfaces, which is usually obtained by extracting point features and then identifying correspondences between the two~\citep{yang2023learning}. Non-rigid approaches take this initial alignment as input and deform the preoperative liver volume or surface, so that it reflects the intraoperative state and reveals the hidden structures to surgeons via AR~\citep{Pfeiffer2020}.  
%However, registration is challenged by factors such as the intraoperative limited field-of-view of the liver surface, intraoperative noise due to camera artifacts and light reflections, the scarcity of distinct texture features, large soft-tissue deformations, unknown material properties and unknown boundary conditions. 
%The surgical domain presents registration algorithms with special challenges.
% V2S-Net, as a baseline method in the field, exhibits good performance for  real-time liver volume registration. 
However, current 3D-3D non-rigid registration methods suffer from various challenges, including high sensitivity to volume and surface noise, struggle with large deformation, and a scarcity of high-quality training datasets. % that minimize domain gaps. 
Therefore, we propose a novel neural network that yields better registration results with higher robustness against substantial noise and different levels of deformation. The network aims to calculate physically accurate liver deformation from partial views of the organ surface while implicitly estimating unknown boundary conditions and tissue properties. %Furthermore, we designed a simulation pipeline for synthetic training and test data generation to tackle the data scarcity issue for registration.
The architecture's full potential is enabled by our creation of synthetic training data that specifically target above challenges and help overcome the data scarcity issue for registration.
Our contributions are fourfold:

% maybe highlight the proposed critical design for the method first before refer to the high accuracy ? %'The network aims to ...' sounds like a discussion?

% if the dataset is highlighted as contribution, might be better to add some arguments of why it is important in the intro part.

% the contribution might be too long. 
\begin{itemize}
    \item We propose the Point-based \textbf{Preoperative to Intraoperative Volume to Surface registration (PIVOTS)}, a novel transformer-based neural network for near real-time deformable liver registration. 
    \item We build upon a \textbf{soft-tissue simulation pipeline for registration pair creation} to mitigate the common issue of lacking training and evaluation data.
    % with random shape generation, physical deformation using a hyperelastic material, and diverse partial noisy surface extraction. 
    % \textcolor{red}{We publish the training dataset containing ~200k organ-like shapes and their deformation. TODO: Do we?}
    %concise?: -> the intraoperative surface extraction with different noise levels is teaching the network its robustness against it
    %We develop a simulation pipeline for surgical scene synthesis, featuring high varieties in soft-tissues and their deformation. The pipeline mitigates the issue of limited surgical data, and facilitates training our method.
    \item We curate a high-quality \textbf{healthy human liver breathing motion (HHLBM)} dataset, which can be used to benchmark registration methods, filling the gap of a lack of real evaluation data.
    % MRI scans of the subjects' livers are taken under inhaled and exhaled statuses in prone and supine positions, captured in multiple sessions. This results in eight scans per subject with liver segmentation masks and manually marked vessel bifurcation landmarks, 
    \item Our \textbf{extensive evaluations} show high registration performance, especially strong robustness against substantial amounts of noise, large deformation, and various amounts of intraoperative visibility.
    
    % in terms of target registration error (TRE) and fiducial registration error (FRE), and 
\end{itemize}

Our paper is organized as follows: we introduce related work in 3D registration in Sec.~\ref{sec:related_work}, followed by the elaboration of the proposed non-rigid neural network and synthetic scene generation in Sec.~\ref{sec:methods}. Then we present various evaluation datasets in Sec.~\ref{eval:datasets} and show results of our experiments: A comparison with SOTA methods (Sec.~\ref{exp:comparison}), detailed studies of sensitivity to different deformation levels (Sec.~\ref{exp:deformation_levels}) and noise levels (Sec.~\ref{exp:noise_levels}), and an intraoperative visiblity experiment (Sec.~\ref{exp:visiblity}), followed by a discussion of strengths and remaining limitations of our method.

% \textcolor{red}{TODO}

\section{Related Work}
\label{sec:related_work}
\notes {
- mention dataset sparsity
	- discuss problem that we don't have matched deformation and material property data \(\rightarrow\) can only study it on synthetic data
}

% maybe move this paragraph to introduction:
% Laparoscopic liver registration requires the fusion of pre- and intraoperative data. Recent work focuses on 3D-3D registration where preoperative models segmented from CT/MRI as the source while intraoperative scenes are reconstructed from laparoscopic videos as the target. The key to solving this registration problem is exploring correspondences between the source and target, however, such exploration is challenging due to the lack of salient features, the high amount of noise, and the elevated discrepancy between modalities from different scanners (MRI/CT and laparoscopic videos). 

Registration algorithms, which are the core component in laparoscopic navigation systems, can be broadly divided into 3D-3D and 3D-2D registration based on the input modalities. State-of-the-art 3D-3D registration methods can be further categorized into two primary types: 1) traditional approaches, and 2) learning-based methods. 

Among traditional registration methods, feature-based methods detect keypoints~\citep{sipiran2011harris,streiff20213d3l,jin2024keypointdetr}, which are described with either local or global point descriptors~\citep{rusu2009fast,guo2013rotational,zhao2022alike}. Explicit correspondences are then established through feature matching~\citep{rusu2008aligning,guo2013rotational,rodola2013scale}, or voting mechanisms for correspondence optimization~\citep{tombari2010object,yang2023mutual}. 

In contrast to methods relying on one-shot correspondence, which are highly dependent on the quality of feature descriptors and computationally expensive, iterative techniques such as ICP~\citep{besl1992icp} and its variants (e.g. Go-ICP~\citep{yang2015go}, non-rigid ICP~\citep{amberg2007optimal}) optimize both point correspondences and transformation. 
%should the above be grouped with the aforementioned feature-based methods in one paragraph? 
Another family of methods relies on implicit correspondences, for instance, CPD~(\citep{myronenko2010point,hirose2022geodesic}) models the source point cloud as the centroids of Gaussians with equal isotropic covariance matrices in a Gaussian Mixture Model. Moreover, these methods are often stuck in local minima given erroneous initial alignment, and focus on surface-to-surface registration, ignoring the volumetric nature and biomechanical properties of an organ. 
Specifically, in the context of liver navigation, a subset of registration methods employs Biomechanical Models for soft-tissue simulation under boundary conditions to infer intraoperative deformation, often using Finite Element Methods~\citep{ozgur2018preoperative,suwelack2014physics,Yang2024BoundaryCB}. 
% Yet, obtaining realistic intraoperative boundary conditions as deformation constraints remains challenging, thereby limiting the generalizability of these approaches. 
In theory, these methods model material behavior in a physically accurate way. However, in practice they usually require many simplifications which negate this advantage. Furthermore, they are relatively difficult to parallelize, often suffer from long run-times and can have difficulties adapting to larger deformations.

In light of the marked success of deep learning-based methods, data-driven 3D registration enables end-to-end frameworks, offering computational efficiency and robustness. According to different representations of 3D point clouds and deformation, these learning-based methods can be divided into grid-based, graph-based and point-based methods.

Point clouds present challenges for neural networks due to their irregularity, unordered structure, and varying density. Grid-based methods address these challenges by regularizing point clouds through implicit representation of shape and deformation. Often, the 3D data is converted to the (truncated) signed distance field (SDF), where each voxel encodes the distance to the closest surface or points, with the signs indicating whether the voxel center is located inside or outside of the target volume. In such a way, this format allows straightforward application of 3D convolution kernels while maintaining information about the surface. 
Methods such as 3DMatch~\citep{zeng20173dmatch} and PerfectMatch~\citep{gojcic2019perfect} convert local patches into a volumetric representation of truncated distance field and employ a 3D convolutional neural network to learn local descriptors. % missing the information how the local descriptors are used further
V2S-Net~\citep{Pfeiffer2020} transforms preoperative liver volume and intraoperative partial surface into SDF and DF, respectively, while Cue-Net~\citep{liu2024leveraging} integrates user inputs into the voxel grid as auxiliary information to guide the registration. Both train 3D CNN for deformation prediction.
Similarly, \citep{chen2024points} focuses on voxels surrounding the liver volume and surfaces, but uses a transformer architecture to learn deformation.
Despite these advances, voxel representation suffers from several limitations, such as reduced flexibility, since all inputs need to be converted to a fixed voxel grid, and high memory consumption, as every voxel is processed regardless of its proximity to the target, and they are prone to aliasing.
By contrast, with the advent of robust backbones for point cloud analysis, learning on point clouds directly becomes increasingly efficient. For example, architectures such as PointNet~\citep{qi2017pointnet}, PointNet++~\citep{PointNetPP2017}, PointTransformer~\citep{PointTransformer2021}, and KPConv~\citep{thomas2019kpconv} facilitate the development of point cloud registration~\citep{yang2023learning,huang2020feature,sarode2019pcrnet,shi2021unsupervised,yew2022regtr,qin2023geotransformer,li2022lepard}.

Harnessing PointNet~\citep{qi2017pointnet} to extract global features, PointNetLK~\citep{aoki2019pointnetlk} modifies the Lucas \& Kanade algorithm with global feature differences for iterative alignment, whereas PCRNet~\citep{sarode2019pcrnet} directly concatenates global features of source and target to regress poses.
More recently, LiverMatch~\citep{yang2023learning}, RegTR~\citep{yew2022regtr}, GeoTransformer~\citep{qin2023geotransformer} and Lepard~\citep{li2022lepard} utilize KPConv~\citep{thomas2019kpconv} for grid-downsampling. Combined with point feature extraction and self- and cross-attention for feature propagation, the final correspondences and registration results are yielded according to various regression designs.
LiverMatch~\citep{yang2023learning} computes correspondences using a confidence matrix and the predicted visibility scores of source points to determine the transformation matrix.
RegTR~\citep{yew2022regtr} directly regresses the transformed point clouds from the conditioned features via dual MLP decoders. GeoTransformer~\citep{qin2023geotransformer} adopts a local-to-global scheme, where transformation candidates are first generated between patches with conditioned features, then final transformation is the optimal candidate on the global point clouds.
Lepard~\citep{li2022lepard} integrates positional encoding into a two-block transformer architecture to incrementally achieve robust alignment, suitable for both rigid and non-rigid cases.
% registration, where the source point cloud is roughly aligned by the first block with transformer and matching, and the final match and transformation are obtained by the second block based on the transformed point clouds. This scheme benefits both rigid and nonrigid registration cases.
In contrast, NDP~\citep{li2022non} employs positional encoding with progressively increasing frequencies to address non-rigid registration by shifting from stiff and global alignment to soft and local adjustment.

Another family of 3D learning is Graph-based methods, beginning with the introduction of graph convolutional networks~\citep{scarselli2008graph} and later incorporating convolutional networks~\cite {duvenaud2015convolutional}. Dynamic Graph CNN (DGCNN)\citep{wang2019dynamic} applies dynamic edge convolution to capture both local and global features in point clouds. 
% \citep{wang2019dynamic} applies dynamic edges on point cloud for both local and global feature learning.
DeepClosestPoint~\citep{wang2019deep} exploits DGCNN to extract local features for correspondence prediction with an attention module, followed by a differentiable SVD layer for transformation calculation.
% DGCNN formula from DCP: xl  i = f ({hl  θ (xl−1  i , xl−1  j ) ∀j ∈ Ni})
Predator~\citep{huang2021predator} first extracts keypoint features using KPConv, then applies dynamic edge convolution and cross-attention to predict an overlap score between the two groups of superpoints, which are decoded into correspondences by a linear decoder. 
GraphSCNet~\citep{qin2023deep} constructs a deformation graph over the source point cloud and employs a graph-based local spatial consistency measure to aggregate and embed correspondences, thereby effectively filtering out outliers for robust non-rigid registration. 
Similarly, GCNNet~\citep{zhu2022neighborhood} utilizes GNN to enhance semantic encoding between local regions combined with geometry encodings constructed from the k-NN for multi-level feature extraction.
However, while these graph-based methods perform well on surface-to-surface registration tasks, no volume-to-surface registration is discussed, which is essential for accurate vessel tree and tumor registration in liver navigation systems.

%A bit abrupt: maybe start with 'Apart from the mainstream 3D-3D approaches,'
Recently, the registration problem in liver navigation has been approached from a 3D-2D formulation where the target is 2D laparoscopic images with hand-crafted or automatically detected features. Given that the segmented source liver model is textureless, only geometrical features can be extracted such as liver ridge, falciform ligament and silhouette based on camera views.
Works such as \citep{robu2018global,koo2017deformable,pmlr-v227-labrunie24a} manage to extract corresponding landmarks on 2D images and minimize the discrepancies between 3D and 2D landmarks, achieving rigid or non-rigid registration. Alternatively, approaches like ~\citep{collins2016robust} enrich the source model with real texture by first aligning intraoperative images with the 3D model, followed by deformable registration based on the deviation in textures and boundaries. The P2ILF challenge~\citep{ali2025objective} benchmarks 3D-2D registration by providing a dataset with manually labeled camera poses, where camera pose optimization with differential rendering demonstrates superior performance. Additionally, \citep{mhiri2025neural} proposed a learning-based method that combines 3D synthetic deformation and 2D landmarks to predict intraoperative deformation, yet only patient-specific.
One downside of the 3D-2D formulation is that a single view often does not capture enough of the organ. \citep{espinel2021using} tries to solve this issue by using multiple views. Nevertheless, other limitations such as the inherent ambiguity between 3D models and 2D images, landmarks, hinder the accurate registration between the two. 
% Therefore, the proposed method is still based on the well-established strategy for liver navigation, i.e. 3D-3D registration, the goal is to further improve the accuracy in terms of intraoperative noise and high occlusion. 
% Therefore, while significant advancements have been made in 3D-2D registration methods, 3D-3D registration remains the prevailing strategy in liver navigation. 

In this work, we explore a 3D-3D registration algorithm that works directly on volume and surface point clouds, runs at very high speeds and aims to further enhance registration accuracy in the presence of intraoperative noise and high occlusion, addressing key limitations of existing approaches.

% At this point, the topic of training data has not really been touched upon. There could be a sentence or two highlighting how much training data which methods (roughly) need, or the fact that many methods can't be directly translated to liver surgery without further work because of the lack of training data. It should be clear at this point that we _need_ to generate training data because there isn't really any other solution. 
% For example, after reading this I don't know how LiverMatch is trained
% I think it's important to show that all of these other methods don't need to build their own training data. The fact that we can generalize from synthetic to real data is a major selling point of our paper, even if V2S-Net was similar - we do it at a never-before-seen scale!!

Notably, most of these methods are not set up for the volume-to-surface registration task, or require retraining for application in the surgical domain. However, no available real or phantom dataset contains enough registration pairs for such training, since 3D-3D registration data is very difficult and time-consuming to obtain. % [I'd like to cite some, but then we're going to be asked why we're not validating on them].
Phantom datasets provide the highest number of samples, with 112 registration pairs in the SparseDataChallenge dataset \citep{Heiselman2024} and six pairs in the private phantom dataset presented in section \ref{eval:datasets}. Three datasets capture in vivo human liver deformation due to breathing: 3D-IRCADb-02 with one subject and one deformation \citep{ircadb-02}, the Breathing Motion dataset with two patients and one deformation \citep{Pfeiffer2020} and the new HHLBM dataset currently comprising seven patients with seven deformed states each \ref{eval:datasets}. Intraoperatively, the DePOLL porcine dataset with 12 deformed states of one organ \citep{modrzejewski_vivo_2019} and a clinical dataset with four patients \citep{rabbani_methodology_2022} are available.
%- Breathing Motion 2 x 1 in vivo
%- HHLBM 3 x 7 in vivo
%- DePOLL 1 x 12 intraoperative (pig)
%- Rabbani2022 4 x ? intraoperative
%- SparseDataChallenge -> initial alignment not shared = not reproducible
%- IRCADb-02 (1 patient)

In addition to the low number of samples, depending on the original aim of the data collection, only landmarks or sparse correspondences may be available, whereas a dense displacement field is needed for training the volume-to-surface registration task. It has been shown that registration networks can successfully transfer from synthetic training data to applications in the real domain~\citep{pfeiffer_learning_2019, Pfeiffer2020,ali2025objective}, %maybe P2ILF?]
which adds the benefit of control over the experimental frame and the possibility to extract any information as needed. Therefore, for this work we also rely on synthetic data for training, generated using the pipeline by Pfeiffer et al. \citep{pipeline_longabstract2025}, and show that its composition allows for generalization with respect to shapes, intraoperative view, deformation magnitude and noise levels during inference.

\begin{figure*}[!ht]
    \centering
    % \frame{
    \includegraphics[trim=0cm 17.3cm 1.7cm 0cm,clip,width=\textwidth]{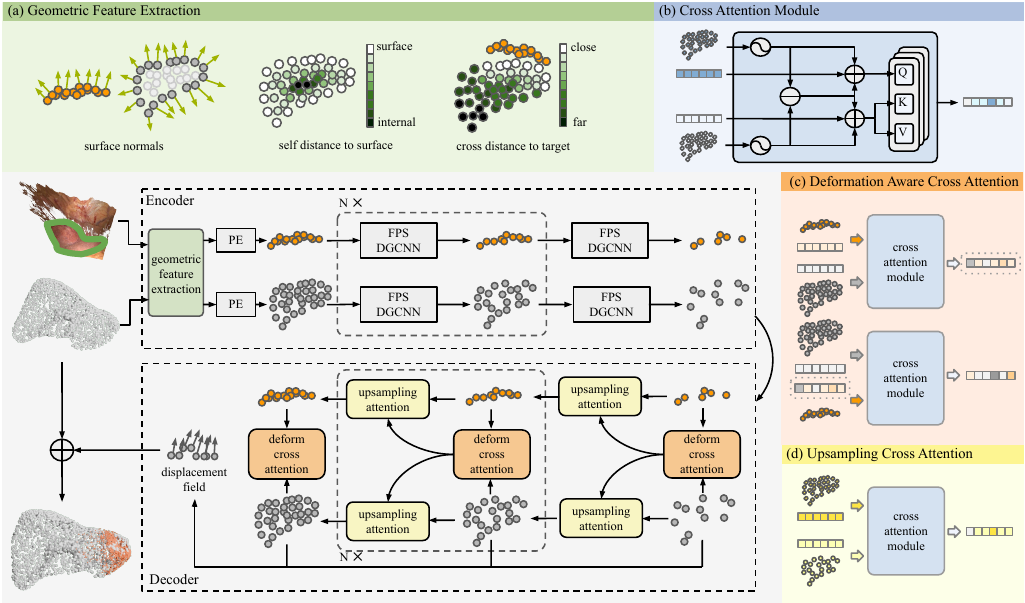}
    % }
    \caption{Architecture of the proposed registration network. (a) For the intraoperative target points as well as the surface points of the preoperative point cloud, we encode the surface normals (see Sec.~\ref{method:geometric}). For the preoperative source points we additionally encode the smallest distance to the mesh surface as well as the distance to the nearest intraoperative surface point. (b) The point attention module is a core component of the network (described in Sec.~\ref{sec:relative_point_attention}). (c) Building upon this attention module, the Deformation Aware Cross Attention combines information from pre- and intraoperative point clouds (see Sec.~\ref{sec:deformation_aware_cross_attention}). (d) The point attention module is also used for the Upsampling Cross Attention module, which propagates features from lower resolutions of the pre- and intraoperative points to higher resolutions (see Sec.~\ref{sec:upsampling_cross_attention}). The final network architecture (left) partially resembles a U-Net, with information first being condensed to a lower resolution and then iteratively refined back to the original resolution.}
    \label{fig:network}
\end{figure*}

\section{Methods}
\label{sec:methods}

Given a preoperative liver volume from MRI / CT, represented by point cloud $\mathbf{V} = \{\mathbf{v}_i \in \mathbb{R}^3 \mid i = 1, \ldots, N\}$, and an intraoperative partial liver surface reconstructed from laparoscopic video, noted as point cloud $\mathbf{S} = \{\mathbf{s}_j \in \mathbb{R}^3 \mid j = 1, \ldots, M\}$, a non-rigid 3D-3D registration algorithm aims to find a transformation $\mathcal{T}: \mathbb{R}^3 \rightarrow \mathbb{R}^3$ that deforms the source $\mathbf{V}$ to align with the fixed target $\mathbf{S}$. The deformed $\tilde{\mathbf{V}}=\mathcal{T}(\mathbf{V})$ should reflect the current status of the intraoperative liver. In our case, such a transformation $\mathcal{T}$ can be represented by a 3D displacement field $\Phi \in \mathbb{R}^{N \times 3}$, i.e. $\mathcal{T}(\mathbf{V})=\mathbf{V}+\Phi$. Our goal is to build a deep network \textbf{PIVOTS} which can predict $\Phi$ for all points in $\mathbf{V}$ (internal and surface points) directly from the geometry $\mathbf{V}$ and $\mathbf{S}$, i.e. 

\begin{equation}
    \Phi = \text{PIVOTS}(\mathbf{V}, \mathbf{S}).
\end{equation}

% Overview:
Our method adopts a dual-stream encoder-decoder architecture. At the encoder branch, we use farthest point sampling~(FPS) and DGCNN~\citep{wang2019dynamic} to iteratively downsample the preoperative and intraoperative point clouds and propagate features at each level. 
The decoder uses these features to predict the displacement field at each level, using cross-attention to merge information from $\mathbf{V}$ and $\mathbf{S}$ and propagate the merged information between layers.
The architecture overview is shown in Fig. \ref{fig:network} and code is available at \url{https://github.com/pengliu-nct/PIVOTS}. The following will describe all components of the architecture in detail.
% more description...

% \subsection{Non-rigid Registration Network}

% \subsubsection*{Overview}

% asterisk for 
% \begin{figure*}[!ht]
%     \centering
%     \includegraphics[trim=0cm 0cm 0cm 0cm,clip,width=\textwidth]{network.png}
%     \caption{Placeholder for network architecture.}
%     \label{fig:network}
% \end{figure*}

% \subsubsection*{Core components}

\subsection{Point Cloud Preprocessor}

The original preoperative and intraoperative point clouds are first enhanced by preprocessing steps before being fed into the registration network for deformation prediction. This serves as a standardization scheme for data from various sources and creates additional geometric information which the network can act upon.

First, both pre- and intraoperative meshes are re-sampled to have a common resolution of roughly 5\,mm, and several pre-computed features are extracted to help the network understand the geometry of inputs. \textit{Point distance field, $F_d$}: for each point in the preoperative volume $\mathbf{V}$, we compute the distance to the closest point in intraoperative surface $S$. Likewise, for each point in $\mathbf{S}$ the smallest distance to the surface of $\mathbf{V}$ is computed. Additionally, for each internal point of $\mathbf{V}$, we compute its distance to the closest point on the surface.
\textit{Surface normals, $F_n$}: for the surface points in $\mathbf{V}$ and $\mathbf{S}$, surface normals are computed to help the network identify the orientation of surfaces. \textit{Sinusoidal positional encoding}: for pre- and intraoperative point clouds, sinusoidal positional encodings are computed with 8 frequencies with the mappings: $\xi: \mathbb{R}^3 \rightarrow \mathbb{R}^{6 \omega}$, e.g., positional encoding of preoperative points is $\xi_v=[sin(2^ \omega v_i), cos(2^ \omega v_i)]$, where $\omega \in (2^{-1},2,4,8,16,32,)$ is the list of encoding frequencies. 

Moreover, to allow batch-processing, both pre- and intraoperative point clouds are further set to exactly $n$ points by either subsampling randomly or by adding dummy points that are way outside the space of interest and ignored during farthest point sampling (FPS). The final inputs to the network are: $\mathbf{V}_r=\{P_v, \xi_v, F_v\}$ and $\mathbf{S}_r = \{P_s, \xi_v, F_s \}$, where $P_v, P_s \in \mathbb{R}^{n \times 3}$, $\xi_s, \xi_s \in \mathbb{R}^{n \times (6 \omega)}$ and $F_v, F_s \in \mathbb{R}^{n \times 5}$ (see Fig. \ref{fig:network} (a)).
We store all distances and locations in meters.

% the hidden story is encoder-decoder but here introduce key components instead...

\subsection{Geometric Feature Extraction} % encoder
\label{method:geometric}
% FPS + DGCNN
Downsampled point clouds no longer retain geometry information compared to original volumes, therefore, we adopt the widely used DGCNN~\citep{wang2019dynamic} to extract and propagate robust local geometry features between $L$ layers in different resolutions obtained by farthest point sampling (FPS): $(P^{\ell+1}, \xi^{\ell+1}, F^{\ell+1}) = \Pi_{fps}(P^\ell, \xi^{\ell}, F^\ell, N^\ell)$, where $\Pi_{fps}$ refers to mapping of FPS function, $N^\ell$ is the number of points at $\ell$-th layer (see a in Fig.~\ref{fig:network}). Features of next layer $F_v^{\ell+1}, F_s^{\ell+1}$ are updated using two-stream edge convolution for $\mathrm{V}_r$ and $\mathrm{S}_r$ separately due to the fact that the geometry information of pre- and intraoperative data is naturally different:

\begin{align}
    % \hat{F}_v^{l+1} = \max_{j \in \mathcal{K}(i)} \mathrm{ReLu}(\Phi(F_{v,j}^l - F_{v,i}^{l+1}, F_{v,i}^{l+1}))
    % \hat{F}^{l+1} = \max_{j \in \mathcal{K}(i)} \mathrm{ReLu}(\Phi(F_{j}^l - F_{i}^{l+1}, F_{i}^{l+1}))
    % \hat{F}^{l+1} = \max_{j \in \mathcal{K}(i)} \mathrm{ReLu}(\mathcal{H}(F_{j}^l - F_{i}^{l+1}, F_{i}^{l+1}))
    \hat{F}^{\ell+1} = \max_{j \in \mathcal{K}(i)} \mathrm{ReLu}(\mathrm{MLP}(F_{j}^\ell - F_{i}^{\ell+1}, F_{i}^{\ell+1}))
\end{align}
where $F^\ell$ is features of either preoperative or intraoperative point clouds, $\mathcal{K}$ represents the K nearst neighors from layer $\ell+1$ of point $i$ in layer $\ell+1$, and features are aggregated by the a multi-layer perception $\mathrm{MLP}$, followed by $\mathrm{ReLu}$ and maxpool layers.

Using such an encoder constructed by DGCNN and FPS, point clouds are downsampled fast, and the local geometry features are extracted and propagated between layers at different resolutions, which will be refined by the decoder layers for deformation prediction.

% Encoder takes input coordinates $P_P \text{and} P_I$ and features $F_P \text{and} F_I$ and performs Farthest Point Sampling (FPS) to obtain points in different resolutions:
% \begin{equation}
%     {P^i_P, P^i_I, F^i_P, F^i_I} = \mathrm{DGCNN}(\mathrm{FPS}(P^{i+1}_P, P^{i+1}_I, F^{i+1}_P, F^{i+1}_I))
% \end{equation}

% \begin{equation}
%     \mathrm{DGCNN} = Conv(concat(F^i_P - F^{i+1}_{P, K}, F^i_P))
% \end{equation}
% where the coordinates and features of the current layer are the sampling results from the last layer. DGCNN is employed to propagate features between layers, taking the features of center points $P^i_P$ and the $K$ neighbors $P^{i+1}_{P, K}$.

\subsection{Biomechanical Deformation-aware Transformer} % decoder

Pre- and intraoperative liver point clouds are orderless and can vary largely in density, shape, and appearance. Thus, it is substantial to design a mechanism that allows features to flow and propagate between the two point cloud pairs in the same or different resolutions. Attention mechanism demonstrates promising performance for point cloud analysis due to its invariance to point order and its ability to enable dynamic information interaction between two arbitrary point clouds. These make the attention mechanism a suitable and efficient analyzer in dealing with pre- and intraoperative point cloud pairs. Therefore, we proposed a novel \textit{Biomechanical Deformation-aware Transformer} consisting of \textit{Deformation Aware Cross Attention} and \textit{Upsampling Cross Attention} modules using \textit{Relative Point Attention} as the core, for pre- and intraoperative point cloud feature interaction and propagation. They are introduced in Sec.~\ref{sec:deformation_aware_cross_attention} and Sec.~\ref{sec:upsampling_cross_attention}.

% Attention mechanism demonstrates promising performance for point cloud analysis due to its invariance to point order and its ability to dynamically capture long-range dependencies without relying on fixed receptive fields. 
% Given input point features $F_X$ and $F_Y$, attention function $Attn$ first generates
% introduce attention machanism in general

% \textbf{Feature Disentangled Point Attention}
\subsubsection{Relative Point Attention}
\label{sec:relative_point_attention}

We propose a \textit{relative point attention} (see b in Fig.~\ref{fig:network}) which is used in multiple parts of our decoder to compute new features of a \emph{query} point $P_Q \in \mathbb{R}^{N \times 3}$ from its local neighborhood in the target point cloud. We employ vector attention starting with finding the k nearest neighbors $P_K \in \mathbb{R}^{N \times k \times 3}$ of $P_Q$, which act as the \emph{key} and \emph{value} points. To each feature vector ($F_Q \in \mathbb{R}^{N \times f}$ and $F_{K,k} \in \mathbb{R}^{N \times k \times f}$), we concatenate the point's coordinates both in absolute form and relative to $P_Q $ and $P_K$, as well as positional encodings $\xi$ (sine and cosines at different frequencies) thereof. We then calculate the query $Q$, key $K$ and value $V$ vectors as:

\begin{align}
    % Q, K, V = MLP(F_P, F_I) \cdot (W_Q, W_K, W_V)
    % Q, K, V = Map(F_P, F_I) = Fp \cdot W_Q, F_I \cdot W_K, F_I \cdot W_V
    % Q = W_q Fp \quad (K, V) = (W_k, W_v) F_I
    Q &= W_Q (\xi_Q \oplus (\xi_Q \ominus \xi_K) \oplus F_Q) \\
    K &= W_K (\xi_Q \oplus (\xi_Q \ominus \xi_K) \oplus F_{K,k}) \\
    V &= W_V (\xi_Q \oplus (\xi_Q \ominus \xi_K) \oplus F_{K,k}) 
\end{align}
where $W_q, W_k, W_v \in \mathbb{R}^{c \times f}$ are linear projection matrices, resulting in $Q \in \mathbb{R} ^ {c \times N}$ and $K, V \in \mathbb{R} ^ {c \times N \times k}$, $\xi \in \mathbb{R}^{6\omega}$ is sinusoidal positional encoding, here we also include the the positional encoding of the $N$ query points to their $k$ neighbors. $F_Q$ is the point cloud features, and $F_{K,k}$ is the features of the k nearest neighbors of query points. $\ominus$ is an operation for point-wise relative position calculation, and $\oplus$ is a concatenation operation. 

Then, attention weights are then calculated using \textit{local additive attention}: 
% The attention process ($F_P$ attends to $F_I$) can be denoted as:

    % Q, K, V = \math{MLP}(F_P, F_I) \cdot (W_Q, W_K, W_V) \\
% \end{align}
\begin{equation}
    % \text{Attn} = \text{Softmax}\left(\frac{Q + K}\right)V
    att(Q, K) = Softmax\left(\mathrm{MLP}(\mathrm{tanh}(Q + K))\right) 
\end{equation}
where $att \in \mathbb{R}^{d \times N \times k}$ is the attention score of $Q$ on $K$. For each key point, the weighted average of the value features $V$ is computed, resulting in the new features $\tilde{F_Q} \in \mathbb{R}^{d \times N}$ for the query point $Q$. The query feature $F_Q$ is then the combination of multi-heads of attention modules:

\begin{equation}
    \tilde{F}_Q \leftarrow \mathrm{MultiHead}(Q, K, V) = \mathrm{MLP}(cat[att_i(Q, K) \cdot V])
\end{equation}
where $\tilde{F}_Q \in \mathbb{R} ^ {N \times d^\prime}$ is the updated query features. In practice, we set the number of attention heads to four.

\subsubsection{Deformation Aware Cross Attention}
\label{sec:deformation_aware_cross_attention}

To predict an accurate alignment between the preoperative and intraoperative point clouds, we must allow information to flow between the two clouds. Therefore, based on the proposed local \textit{relative point attention} module, we assemble a \textit{deformation aware cross attention} module (see c in Fig.~\ref{fig:network}) between preoperative volume points $\mathbf{V}^\ell_r = (P_v^\ell, \xi_v^\ell, F_v^\ell)$ and intraoperative surface points $\mathbf{S}^\ell_r =(P_s^\ell, \xi_s^\ell, F_s^\ell)$ at each resolution $\ell$ to extract features that are ready to be regressed to predict displacement fields.

The module consists of two cross attention steps. To gain rich information about preoperative points for all intraoperative points, we first employ $\mathbf{S}^\ell_r$ to attend to $\mathbf{V}^\ell_r$:

\begin{align}
    \tilde{F}_s^\ell & \leftarrow \mathrm{MultiHead} ( Q_s, K_v, V_v ) 
\end{align}
where $Q_s$ is the linearly transformed surface features $F_s^\ell$ and $K_v, V_v$ come from volume features $F_v^\ell$. 

With the updated intraoperative surface features being informative of preoperative data, the second attention module is applied, aiming to convey such knowledge to the preoperative point:

\begin{equation}
    \tilde{F}_v^\ell \leftarrow \mathrm{MultiHead} ( Q_v, \tilde{K}_s, \tilde{V}_s ) 
\end{equation}
where $\tilde{K}_s, \tilde{V}_s$ stem from the updated surface features $\tilde{F}_s^\ell$, and $\tilde{F}_v^\ell$ is the updated preoperative features, being aware of the intraoperative points at the current resolution. At this point, information can flow back and forth between the point clouds, and can propagate between neighborhoods, and $\textit{F}_P^\ell$ should contain all necessary information about deformation. 
% i.e., visible points know where to deform, and the points without corresponding points in the intraoperative points can obtain knowledge from visible points and know where they can move.
Hence, we let the network predict a (downsampled) version of the displacement field $\Phi^\ell$ for every level $\ell$:

\begin{equation}
    \Phi^\ell = \mathrm{MLP}(\mathrm{ReLu}(F_P))
\end{equation}

Such a design enables the information flow in terms of larger deformation between pre- and intraoperative points with the \textit{local additive attention}, as all points from both sides gain the opportunity to update their knowledge regarding their counterparts. We demonstrate the effectiveness in the ablation study in Sec.~\ref{sec:ablation}.
% \textbf{Decoder} is responsible for predicting deformation at each resolution and propagating features from lower layers to higher layers. To predict deformation at layer $i$, we perform cross-attention where intraoperative features first query preoperative then the other way around:

% \begin{equation}
%     F_I^P = MLP ( Attn(F_I, F_P))
% \end{equation}

% \begin{equation}
%      F_P^I = MLP ( Attn(F_P, F_I^P))
% \end{equation}

% Such cross-attention enlarges the receptive fields which are potentially narrowed by only looking at the neighbors during vector attention. The conditioned preoperative features are regressed to displacement field $\phi_i$ and propagated to the next layer using the same attention mechanisms. The output of the decoder is a set of such displacement fields of $L$ layers: $\{\phi_i|i=0,...L\}$, and the loss function is the weighted sum of the mean square errors (MSE) at each resolution:

\subsubsection{Upsampling Cross Attention}
\label{sec:upsampling_cross_attention}

The abstract features calculated at lower resolutions must later be incorporated back into the features for the higher resolutions in the decoder. For this, \textit{Upsampling Cross Attention} modules (see d in Fig.~\ref{fig:network}) are built to propagate information from lower layers to higher layers. Taking preoperative volume as an example:

\begin{equation}
    \tilde{F}^{\ell - 1}_v \leftarrow \mathrm{MLP}(Attn(F^{\ell-1}_v, F^\ell_v))
\end{equation}
where $F^\ell_v$ (the feature vector at the $l$-th layer in lower resolution) is used as the key and value vectors and $F^{\ell-1}_v$ (the feature of ($\ell$-1)-th layer in higher resolution) is used as the query, which will be updated by cross attention to be $\tilde{F}^{\ell - 1}_v$. Similar operations are applied between intraoperative points.

% \notes{
% % - downsampling branch with self attention
% - downsampling branch with DGCNN
% - upsampling branch with cross attention
% }

\subsection{Supervision}

At each decoding layer, a displacement field $\Phi_\ell$ is predicted at each layer $\ell$, so the total outputs of an $L$ layers network is:

\begin{equation}
    \Phi^L = \{ \Phi^\ell \in \mathbb{R}^{N \times 3} | \ell=1, ..., L \}
\end{equation}
where $\Phi^L$ is the predicted displacement field at the highest resolution.

We train the network in a supervised learning fashion on a synthetic dataset generated using a simulation pipeline (see section ~\ref{method:simulation_pipeline}). For each training sample, ground truth displacement field $\Phi_{gt}$ is known, so the predicted displacement fields can be compared at each respective resolution using mean squared error (MSE). The total loss function $\mathcal{L}$ is computed as a weighted sum of the losses:

\begin{align}
    \mathcal{L}_{MSE} & = \frac{1}{N} \sum_{i}^{N} \left\lVert \Phi_{gt,i} - \Phi_i\right\lVert^2_2 \\
    \mathcal{L} &= \sum_{\ell}w_\ell \mathcal{L}_{MSE,\ell}    
\end{align}
where $\Phi_{gt,i}, \Phi_i \in \mathbb{R}^3$ are the predicted and ground truth displacement vectors of a preoperative point. The lower-level losses guide the network to output better features for displacement estimation at high resolution layers, speeding up convergence, while higher-level losses ensure the fidelity and accuracy of the final prediction \citep{Pfeiffer2020}.

% commissioned: 300k, valid from pipeline side: 64,645 + 64,396 + 64,522 = 193,563

\begin{figure*}[!ht]
    \centering
    \includegraphics[width=0.9\textwidth]{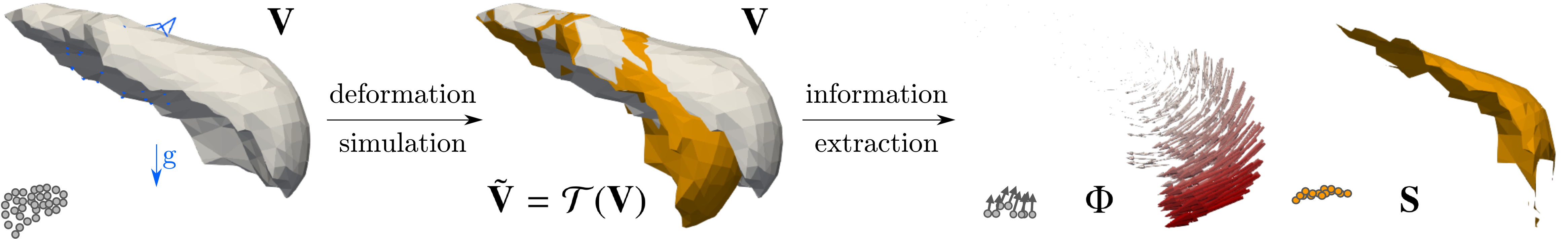}
    \caption{Training data generation. From left to right: Random boundary conditions (blue), including gravity, fixed boundaries (dots) and linear spring boundaries (lines), are placed around a randomly generated organ-like shape with random material properties, representing the preoperative volume $\mathbf{V}$. Simulation under these conditions leads to a deformed configuration $\tilde{\mathbf{V}}$ (orange) which models the deformation encountered during surgery, e.g. due to tool-tissue interaction. Finally, the ground truth dense displacement field $\Phi$ (magnitude-coloured glyphs) and a partial surface with added noise (orange), mimicking an intraoperative point cloud $\mathbf{S}$ reconstructed from the laparoscopic video, can be extracted as training inputs.}
    % perfectionism: find sample with nice big deformation, use visual ligament instead of springs, improve fixed bc
    \label{fig:training_pipeline}
\end{figure*}

\subsection{Synthetic Training Data Generation}
\label{method:simulation_pipeline}
The synthetic training data generation builds upon a simulation pipeline by ~\citep{pipeline_longabstract2025}. Our training data is designed towards providing the network with two core abilities: i) generalizing to new patients without retraining and ii) performing robustly in challenging intraoperative conditions.

Generalization is fostered with respect to organ shape, material and boundary conditions. Diverse, randomly generated organ-like shapes support the adaptation to the natural variation of patients' liver shapes. We try to address the unknown highly variable tissue properties by covering the full range of tissue material parameters. In order to approximate the unknown intraoperative forces which are extremely difficult to measure, like the effects of the pneumoperitoneum, we create a variety of scenarios with differences in magnitude as well as position and surface area which the initial forces act on.

The challenging intraoperative conditions are abstracted as partial views and unavoidable point cloud reconstruction noise. Since it is difficult to encompassingly characterize the different noise properties, we emulate it with a mixture of two noise types in varying strength such that the network can learn to mitigate their effects.
% leaving CT slicing noise out here since we don't apply noise to the preop volume, let's not wake sleeping dragons

The following sections provide more detail about the training data generation. It starts with the scene setup, implementing the shape variability and placement of boundary conditions, followed by simulating the material's behaviour under the specified forces, extracting a part of the deformed surface and adding noise.

\subsubsection{Shape variability} % SHAPES Towards generalization to new patients: 
A scene comprises of a deformable object and its surroundings. The liver is represented by a random organ-like shape, created according to a previously published algorithm~\citep{pfeiffer_non-rigid_2022}: A sphere is subjected to several extrusions and subtractions, followed by smoothing.  Convex parts reflect the anterior side while concave parts mimic the concavities around the hepatic portal on the posterior side. This results in a shape of roughly the dimensions \(d_x=d_y=d_z \sim U(100, 300)\)\,mm, an example is illustrated in figure \ref{fig:training_pipeline}. Figure 1 %todob adapt when supplementary changes
in the supplementary material shows the variability of the created shapes alongside two exemplary real livers from the AMOS dataset~\citep{ji2022amos}, which display the natural variation. In order to create space within the synthetic abdominal cavity, \(0\) to \(3\) more organs of the size \(d_x=d_y=d_z \sim U(50, 200)\)\,mm are created additionally. A convex hull spans around these objects with a minimum distance according to \(\sim U(0, 50)\)\,mm, expanded up to $20$\,mm of Perlin noise along its outward normal, corresponding to a simplified abdominal wall. One or two random paths of up to $100$\,mm along the deformable organ and extruded towards the abdominal wall imitate ligaments. %to find matching surface points to connect the two scene elements. 
Finally, a fixed boundary is placed by expanding from a random face of the deformable organ until \(2\) to \(5\)\% of the surface is covered. 
A comprehensive list of parameters for the scene generation can be found in the supplementary material.

\subsubsection{Deformation and dense displacement field} % BC PROPERTIES DENSE
% I don't go into depth about the actual details of the scene objects, their references and the pipeline blocks here because I think that would be too implementation-oriented
Subsequently, the created scene objects are equipped with physical properties. Trying to mimic liver tissue, the deformable organ is modelled as a hyperelastic and nonlinear material with a Young's modulus \(E \sim U(3, 30)\)\,kPa and a Poisson's ratio \(\nu \sim U(0.45, 0.48)\) informed by literature values \citep{glozman_method_2010, mattei_sample_2016, marchesseau_nonlinear_2017, estermann_hyperelastic_2020, lemine_mechanical_2024}, sampled independently in order to cover the parameter space when creating a sufficiently high number of samples. Hyperelasticity is realized using the isotropic compressible Neohookean formulation by Bonet and Wood \citep{bonet_nonlinear_2008} in SOFA's implementation:
\begin{align*}
    \Psi = \frac{\mu}{2}(I_1 - 3) - \mu \ln J + \frac{\lambda}{2} (\ln J)^2 \\
    \text{with} \quad \mu = \frac{E}{2 (1 + \nu)}, \text{ }\lambda = \frac{E}{3 (1 - 2 \nu)},
\end{align*}

where $\Psi$ denotes the strain energy, $I_1$ the first invariant of the right Cauchy-Green tensor and $J$ the Jacobian. For simplification, the material is assumed to be homogeneous with a mass density of \(\rho \sim U(1.05,1.09)\),g/cm\(^3\)~\citep{lemine_mechanical_2024}.

The previously created surface is processed into a volumetric tetrahedral mesh using GMSH~\citep{gmsh_2009} (mesh generation) and the built-in Netgen interface (optimization) \citep{netgen}, symbolizing the still undeformed, preoperative volume \(\mathbf{V}\).
A maximum mesh element size of 1\,cm presents a trade-off between accuracy and runtime.

In order to create diverse deformation scenarios we employ three kinds of boundary conditions: Gravity, fixed boundaries and linear elastic springs (blue elements on the left of figure \ref{fig:training_pipeline}). The positions of the latter two have been determined in the scene generation step. The fixed boundaries are enforced as a Dirichlet zero-displacement boundary constraint on the deformable organ's surface. The spring boundaries between the abdominal wall and the deformable organ's surface can only move on the organ side. Defined in a state outside of equilibrium, they serve two purposes: In a stretched state, they will pull on the deformable organ, reflecting anatomical ligaments of different strength. In a compressed state, they represent a very simplified version of other organs pushing against the liver.  %The attachment points to the abdominal wall are considered fixed boundaries. 
The springs' initial state is determined by their current length \(l = l_0 + \delta l\), the rest length factor \(c_0= \frac{l_0}{l} \sim U(0.9, 1.1)\) to show the deviation from the relaxed state, and the stiffness \(k \sim U(100, 300) \frac{\text{N}}{\text{m}}, k=\text{const.}\). %\(k\) stays constant over the course of the simulation. 
Each spring \(i\) in a 10\,cm ligament object \(lig\) is assigned the same \(k_{lig}\) and \(c_{0,lig}\), with different \(\delta l_{lig,i}\) based on the slightly different \(l\) due to the initial placement. Each spring's initial force corresponds to \(F_{lig,i} = - k_{lig} \cdot \delta l_{lig,i}\). It follows that \(c_0 < 1\) corresponds to a stretched spring (pulling), \(c_0 > 1\) to a compressed string (pushing).

The simulation is set up using the SOFA framework~\citep{Sofa20212} with an implicit Euler time stepping scheme (step size $50$\,ms) and a Conjugate Gradient linear system solver. %From the initial conditions, 
The simulation is run until equilibrium is reached, but in order to obtain higher deformations for training the network, we export the timestep with the highest deformation instead of the final state, therefore including transient dynamic states of the tissue in the training. The difference between the undeformed state \(\mathbf{V}\) (grey) and deformed state $\tilde{\mathbf{V}}=\mathcal{T}(\mathbf{V}) = \mathbf{V}+\Phi$ (orange) is illustrated in figure \ref{fig:training_pipeline}, second from the left. Since point correspondences between the undeformed and the deformed mesh stay intact, the ground truth dense displacement field $\Phi$ is easily computed, as shown in red colour and magnitude scaled glyphs in figure \ref{fig:training_pipeline}, second from the right.

\subsubsection{Intraoperative surfaces} % NOISE Towards robust intraoperative behaviour: Varying noise levels
In order to recreate the partial and noisy information of the intraoperatively reconstructed point cloud, a part of the surface of the deformed mesh is extracted and perturbed at random~\citep{pfeiffer_non-rigid_2022}.  %following the algorithm described in M. Pfeiffer's work. 
Starting from a randomly selected mesh vertex as a reference, a randomly weighted sum of the geodesic distance, the angle between the normal vectors and a Poisson noise value is used to select the points to remain.  %- point indices are kept consistent with full surface such that correspondence is known
After increasing the number of points on this partial surface by subdivision, an offset is sampled for each direction independently from separate Perlin noise distributions sharing an amplitude $A_P$, using the point positions as an input \(\varepsilon_{x_i} \sim A_P \cdot\text{Perlin}(x_1 \cdot f_{x_i}, x_2 \cdot f_{x_i}, x_3 \cdot f_{x_i}), i \in {1,2,3}\). Additionally, a Gaussian noise contribution is sampled from \(\mathcal{N}(x_i, \sigma) \text{ for } i \in \{1,2,3\}\) and added to the coordinates, followed by a randomized sparsification step to remove few larger areas rather than many small ones. The result is an emulated noisy partial surface $\mathbf{S}$, visualized in orange in figure \ref{fig:training_pipeline} (right).

In this manner, after discarding invalid configurations, we create 193,563 samples for training. A full overview of the parameters used in the data generation pipeline can be found in Table 1 %Todob adapt if changes
in the Supplementary Material.
%- TODO: Name the training dataet here?

\subsection{Implementation and Training}

PIVOTS is implemented in Pytorch and trained on an Nvidia A100 GPU with batch size 20 with a one-cycle learning-rate scheduler \citep{Superconvergence2019} and AdamW optimizer. We use Optuna \citep{Optuna019} to fine-tune our hyperparameters with a smaller subsection of the training dataset and then train with the full training dataset and the most promising parameters for 100 epochs. We use $k=30$ for all neighborhoods in the point attention modules, $e=29$ for all attention embedding vectors, and all input point clouds are standardized to contain $2,500$ points. We build PIVOTS with six down- and upsampling layers ($L=6$) with $[8, 35, 92, 144, 239, 321]$ points for each downsampled layer respectively, and point feature vectors of length [200, 150, 110, 80, 60, 50] during the downsampling and upsampling stages of the network.

\section{Evaluation}
\label{sec:evaluation}
% \citep{Brunet2019}

We evaluate the proposed network on four datasets, including two synthetic datasets, a phantom dataset, and a real liver dataset, which are introduced in Sec.~\ref{eval:datasets}. Extensive experiments are conducted to assess the generalizability to various data sources (see Sec.~\ref{exp:comparison}), robustness against noise (see Sec.~\ref{exp:noise_levels}) and varying amounts of surface visibility~(see Sec.~\ref{exp:visiblity}). Furthermore, qualitative results are compared to those of many baseline methods~(see Sec.~\ref{exp:qualitative}), highlighting the registration qualities of our method.

\subsection{Datasets}
\label{eval:datasets}

% \notes{
% - mention that we don't have GT for a lot of things \(\rightarrow\) our experiments are the best replacement we can do with available data
% }

\begin{table*}[h!]
\centering
\caption{Comparison of evaluation datasets}
\renewcommand{\arraystretch}{1.5}
\begin{tabular}{>{\raggedright\arraybackslash}m{2.0cm} 
                >{\centering\arraybackslash}m{2.5cm} 
                >{\centering\arraybackslash}m{2.5cm} 
                >{\centering\arraybackslash}m{2.5cm} 
                >{\centering\arraybackslash}m{2.5cm}
                >{\centering\arraybackslash}m{2.5cm}}
\hline
 & \textbf{Synthetic} & \textbf{AMOS} & \textbf{Phantom} & \textbf{HHLBM} & \textbf{Laparoscopic} \\
\hline
\textbf{\parbox[c]{3.2cm}{
    Visualization\\
    \footnotesize 
    \tikz[baseline=-0.5ex]\draw[fill=gray,draw=gray] (0,0) circle (4pt); preoperative \\
    \tikz[baseline=-0.5ex]\draw[fill=orange,draw=orange] (0,0) circle (4pt); intraoperative \\
    \tikz[baseline=-0.5ex]{
        \draw[fill=yellow,draw=yellow] (0,0) circle (4pt);
        \draw[fill=cyan,draw=cyan] (0.2,0) circle (4pt);
    } landmarks
}} &
\includegraphics[width=2.5cm]{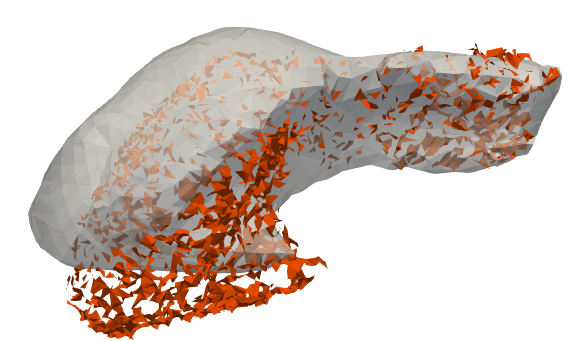} & 
\includegraphics[width=2.5cm]{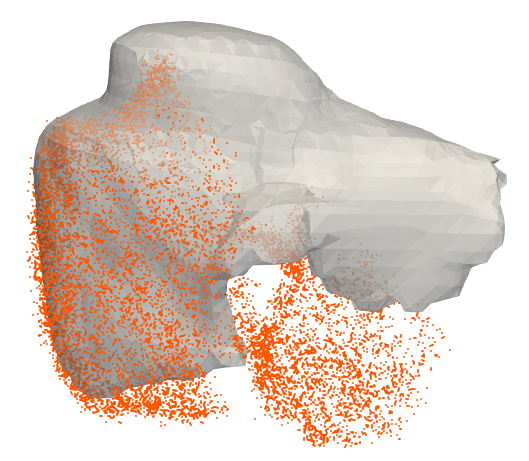} & 
\includegraphics[width=2.5cm]{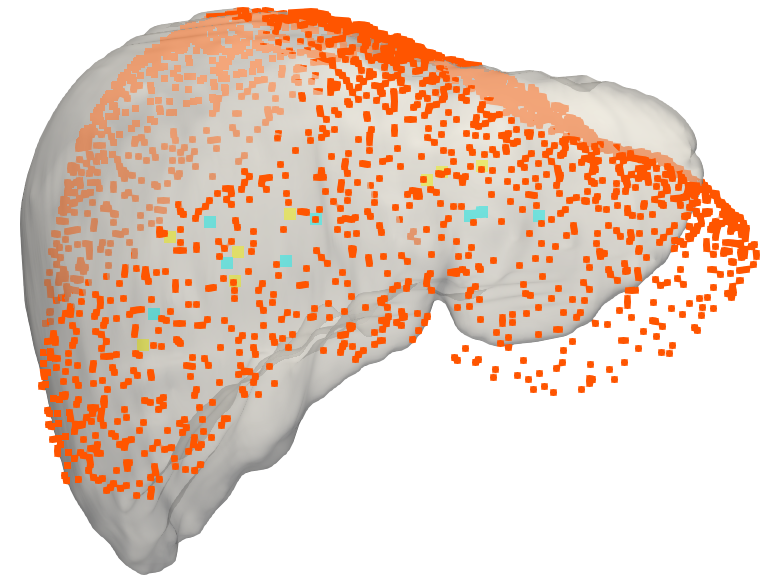} & 
\includegraphics[width=2.5cm]{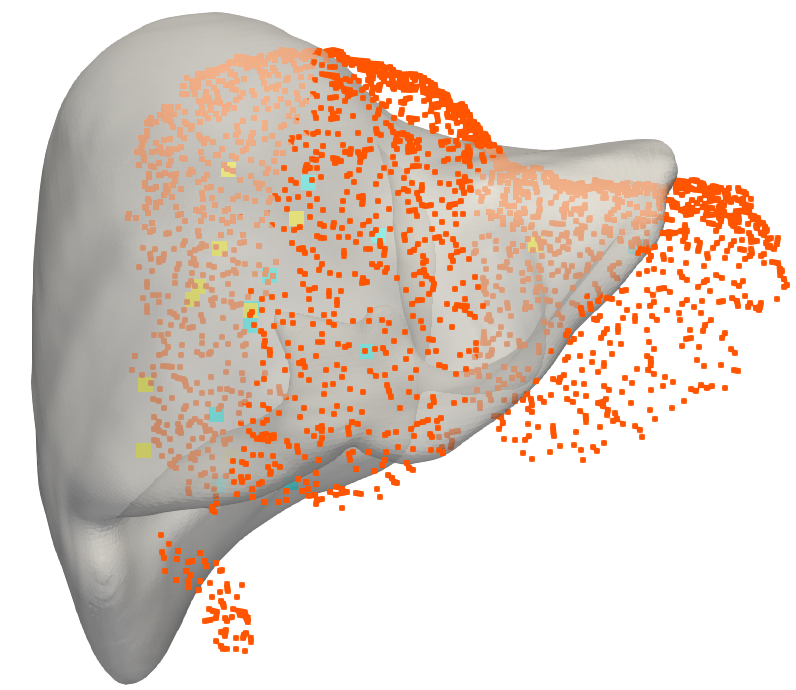} &
\includegraphics[width=2.5cm]{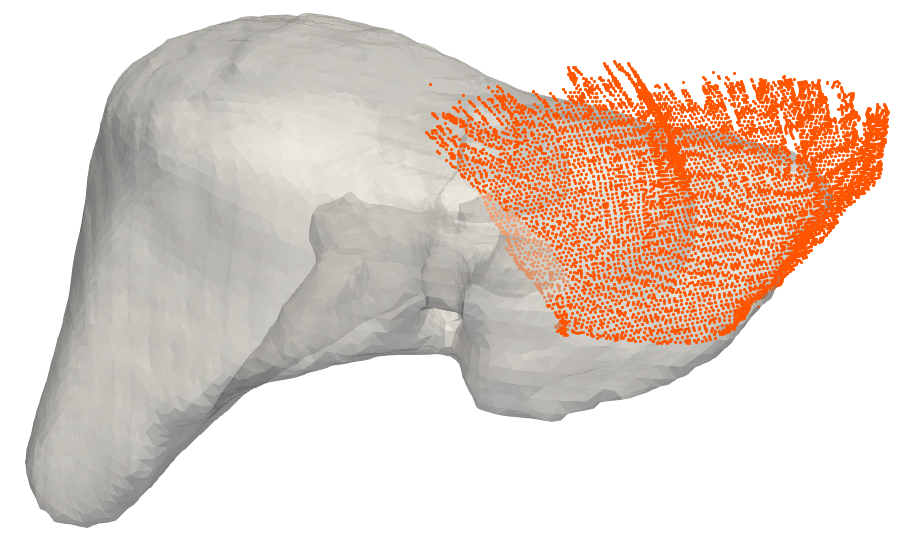} \\
\hline
\textbf{size} & 3008 & 690 ($\times$18 noise) & 6 ($\times$20 targets) & 38 ($\times$20 targets) & 6 \\
% \hline
\textbf{shape} & synthetic & real & phantom & real  & real\\
% \hline
% \textbf{deformation} & synthetic & synthetic & real (tool interaction) & real (breathing, patient positioning)  & real (pneumo\-peritoneum, ligaments, breathing, heartbeat)\\
\textbf{deformation} & synthetic & synthetic & real (tool interaction) & real (breathing, positioning)  & real (pneumo\-peritoneum, breathing, heartbeat)\\
% \textbf{noise} & Perlin and Gaussian noise & Perlin and Gaussian noise & surface and landmarks noise & segmentation and artifacts \\
% \hline
\textbf{metrics} & MED & MED & TRE & TRE  & -\\
% \hline
% 32.14 PRD: 32.04 PRD: 17.35 PRD: 25.78
\textbf{PRD (mm)} & 32.14 & 32.04 & 17.35 & 25.78  & -\\
\hline
\end{tabular}
\label{tab:datasets}
\end{table*}

% \begin{figure}[!ht]
%       \includegraphics[width=\linewidth]{us_phantom_cropped.png}
%     \caption{Phantom setup before MRI acquisition. The rail allows to fix the fake US probe and maintain the deformation during the scan.
%     }
% \label{fig:phantom}
% \end{figure}

The \textbf{Synthetic} evaluation dataset is generated using the same simulation pipeline described in sec.~\ref{method:simulation_pipeline} with the same hyperparameters, resulting in a similar deformation and noise distribution, but the samples are unseen during training. This dataset contains $3,008$ samples. % commissioned: 5000 samples -> only 1561 + 1593 = 3154 valid  from pipeline side

The \textbf{AMOS} evaluation dataset is a semi-synthetic dataset, where the liver models are reconstructed from labelled real abdominal organ CT scans from the AMOS dataset~\citep{ji2022amos} (for details on the data we refer the reader to the original publication) and the deformation is synthetically produced by the same simulation pipeline. $360$ livers are selected, which corresponds to the number of scans with publicly available labels, and $3$ different sets of boundary conditions are applied to each of the real liver models. Randomly selected noise levels are applied to the random partial surfaces as described above. % Todob: Fiona's feedback: metadata about patients (male/female, age, BMI, ...) from AMOS table

For a thorough evaluation of our method, we additionally create a noise level benchmark on this data, AMOS\(_{\text{noise}}\). For this, in order to imitate a real intraoperative surface more closely, we change the partial surface extraction method of the data generation pipeline to mimic a laparoscopic camera view, and fix a set of constant noise amplitude combinations.
The camera view is created by selecting a random position for the camera that remains outside the mesh with a distance of 50 to 200\,mm throughout the simulation, setting its focus to a random point of the point cloud and extracting the whole part of the surface that lies within the field of view.
The selected fixed noise levels include Perlin noise ranging from \(A_P = 0\text{ mm}\) to \(A_P = 15\text{ mm}\) with steps of \(3\text{ mm}\) and Gaussian noise from \(\sigma = 0 \text{ mm}\) to \(\sigma = 5 \text{ mm}\) with steps of \(2.5 \text{ mm}\). Figure \ref{fig:noise_levels} illustrates their effects on the intraoperative surface.
This results in $683$ samples, each one containing $19$ synthetic intraoperative partial surfaces: a random surface with randomly selected noise and one camera view surface for each of the 18 fixed noise levels. % 1080 samples -> only 683 valid from pipeline side (do we additionally exclude any with a too little number of points?)
An overview of the parameters used to generate the AMOS datasets can be found in Table 2 %Todob adapt if changes
in the Supplementary Material.

%\begin{figure*}[!ht]
%    \centering
%    \includegraphics[width=0.8\textwidth]{pics/noise_levels_draft0.png}
%    \caption{\textcolor{red}{Placeholder for exemplary noise level viz.} Left: Full (grey) and partial intraoperative surface (orange) and corresponding partial point cloud. Right: intraoperative partial point cloud with applied noise (red), top row: Gaussian noise only, center row: Perlin noise only, bottom row: combination. Noise levels increase from left to right. ToDo: increase point size and zoom out to make more compact, lighter intraop pc colour, all noise levels same colour, nice formula symbols.}
%    \label{fig:noise_levels}
%\end{figure*}

The \textbf{Phantom} dataset is acquired from a Kyoto Kagaku abdominal phantom \citep{48822b71927c4bf699f7a441c1dede8b} (see figure in supplementary) for controlled experiments on real data with complete knowledge of internal deformation.
A series of six deformations were applied to the liver using a fake MRI-compatible intraoperative US probe. For each compressed state, an MRI scan was performed (plus one for the uncompressed state). In each case a T2 image was acquired with TE=75\,ms, TR=5.2\,s, voxel size=0.88×0.88×3\,mm$^3$ on Philips Ingenia 3.0\,T. The phantom was filled with water to avoid artifacts during scanning. To keep the US probe in a fixed position during MRI acquisition, a specialized construction was designed consisting of a rail, fake probe and a holder (see figure in supplementary). All scans were co-registered rigidly via the basin.
We marked the same eight landmarks at clearly identifiable positions inside the liver scans in 3D Slicer, using the vessel bifurcations as main guidance.
After segmentation of the MRI scans, a partial view of the anterior side of the liver was extracted for each of the six compressed states by simulating the field-of-view of a simple camera. We increased this field of view iteratively (resulting in 20 partial surfaces for each compressed state) and used the uncompressed volume as the preoperative. 
% The target registration errors obtained on the landmarks after applying P-V2S-Net are shown in Fig.~XXX and a sample is shown in Fig.~XXX.

%The acquired images were rigidly co-registered in 3D Slicer based on the phantom container. Additionally, segmentation of vessel structures as well as liver contour were manually performed in 3D Slicer.

%Next, US was acquired using an intraoperative US probe replicating the deformations set during scanning. Vessel cross-sections were then delineated in the US images. 

% from V2S-net paper:
% During breathing motion, the human liver moves and deforms considerably. To assess whether our network translates to real human liver deformation, we evaluate on human breathing data. We extract liver meshes from two CT scans (one showing an inhaled and one an exhaled state) and let these represent the preoperative volume VP and intraoperative surface SI . We search for clearly visible vessel bifurcations and mark their positions in both scans.

% Todo: what is the weighting used?
The \textbf{Healthy Human Liver Breathing Motion (HHLBM)} dataset contains real in vivo liver deformation of healthy subjects during breathing without external disturbance. It enables the evaluation of the proposed network on real human liver deformation. For each healthy subject, data were collected by an MRI scan under three controlled variables: respiratory state (inhaled or exhaled), body posture (supine or prone), and acquisition day (first or second), resulting in a total of eight scan combinations per subject. Images were acquired using a VIBE sequence (T1) with multiple echo acquisitions according to the Dixon method on a Siemens Biograph mMr ($3.0$\,T), voxel size $1.39$x$1.39$x$1.5$\,mm$^3$. Three annotators are involved in the segmentation process, which consist of one biomedical engineer and two experienced computer scientists. To facilitate the time-consuming manual segmentation process, nnUnetV2 \citep{isensee2021nnu} trained on five public liver datasets (LiverHccSeg, Atlas, AMOS, TotalSegmentator MRI, and CHAOS) generates pre-segmentation masks, where only T1-Weighted scans are included to correspond to the focus of the HHLBM dataset. The masks are further refined by the annotators to be the ground truth liver volume. 20 partial surfaces are then extracted from the volumes using the same strategy as for the phantom dataset to reflect intraoperative observations.
For landmarks independent of the surface, since no dense correspondence information is available, two annotators search for clearly visible vessel bifurcations and mark their positions in the eight scans. The mean position of the vessel bifurcations is considered the ground truth position, and the mean deviation between landmarks from the two annotators is $2.76$\,mm. 
Scans with low quality are excluded from the experiments, such as strong artifacts due to the subject moving during the scanning process. 
For evaluating registration methods, one scan from each patient, the inhaled state in the prone position on the first acquisition day, was designated as the "preoperative" volume $\textbf{V}$. The partial surfaces from the other scans of the same patient were chosen as the "intraoperative" partial surfaces $\textbf{S}$, the displacement vectors of the corresponding internal landmarks act as a surrogate for the internal displacement field. In total, data from seven subjects was included in the dataset. Overall, this resulted in 38 valid samples for this dataset.
The data was collected after obtaining ethics approval (TU Dresden, approval number BO-EK-174042023) and with informed consent from every subject.
% Two CT scans are collected from inhaled and exhaled states respectively, and liver meshes are segmented consequently. 

The \textbf{Laparoscopic} Liver dataset comprises six videos collected from real surgery, i.e., six intraoperative surfaces. The preoperative liver models are segmented from CT scans, and the intraoperative 3D scene is reconstructed from a stereoscopic video recorded from an endoscope during the exploration phase at the beginning of a surgery. Since fully automatic 3D reconstruction methods still struggle with the difficult introperative lighting conditions, texture-less surfaces, breathing and heartbeat motion as well as limited visibility, a semi-automatic approach was used to create the intraoperative target surfaces. First, the FoundationStereo method \cite{wen2025stereo} was used for stereo-reconstruction. From the result, overlapping point clouds were hand-picked and aligned using manually placed point correspondences in the CloudCompare Software. Afterwards, points not belonging to the liver surface were discarded. The result is a point cloud containing a partial view of the intraoperative liver, containing noise from the stereo reconstruction as well as alignment errors due to deformation (See Fig. \ref{fig:qualitative}).
% and \textcolor{red}{Supplementary material TODO}).
Patient consent was obtained and the data was collected after approval by the ethics committee (TU Dresden, BO-EK-137042018). %BO-EK-174042023 Bearbeitungsnummer der Ethikkommission for HHLBM https://drks.de/search/de/trial/DRKS00032907/details

% more descriptions here:

\subsection{Evaluation Metrics}

Based on different characteristics of the test sets, we adopt different evaluation metrics. For synthetic and AMOS datasets, ground truth displacement fields are available. Therefore, the mean Euclidean distance (MED) and the root mean squared error (RMSE) are calculated as the dense registration error metric:

\begin{align}
    MED & = \frac{1}{N} \sum \left\lVert \Phi_{pred} - \Phi_{gt} \right\lVert_2 \\
    RMSE & = \left\lVert \frac{1}{N} \sum (\Phi_{pred} - \Phi_{gt}) \right\lVert_2
\end{align}

In contrast, phantom and HHLBM datasets, no dense correspondence information is available, so we employ target registration errors (TRE) to evaluate the model performance:

\begin{equation}
    TRE = \frac{1}{N} \sum \left\lVert \mathcal{T}({L_{P}}) - L_{I} \right\lVert_2
\end{equation}
where $L_{p}$ and $L_{I}$ are the sparse landmarks in pre- and intraoperative liver volume, $\mathcal{T}(L_{P})$ denotes the deformed preoperative landmarks by the estimated displacement field. TRE presents the deviation between the predicted position and actual positions of intraoperative landmarks.

% \subsubsection*{Quantitative Synthetic Data}
% - AMOS: does "realism" of reconstructed shapes have to be justified? (smoothing)

\begin{table*}[htbp]
\caption{Registration performance comparison between baseline methods across different datasets. Bold face represents the best mean registration error on each dataset, and underlined numbers stand for the second best. If a method increases the registration error compared to the dataset's PRD, we consider it failed, indicated by a grey font.}
\label{tab:registration_performance}
\centering
\begin{tabularx}{\textwidth}{@{} Y Z Z Z Z @{}}
\toprule
\multirow{2}{*}{\textbf{Method}} & 
\multicolumn{4}{c}{\textbf{Registration Error (mm, Mean ± SD)}} \\
\cmidrule(lr){2-5}
& \thead{Synthetic} & \thead{AMOS} & \thead{Phantom}  & \thead{HHLBM} \\
& Samples: 3008 & Samples: 683 & Samples: 6 & Samples: 38 \\
& PRD: 32.14 & PRD: 32.04 & PRD: 17.35 & PRD: 25.78 \\
\midrule

% Rigid Registration Group
% \multicolumn{4}{@{}l}{\textbf{Rigid Registration}} \\
% \cmidrule(l){1-4}
% \llap{\multirow{7}{*}{\rotatebox{90}{Rigid Registration}}} 
% \textcolor{gray}{± 23.77}
ICP    & \textcolor{gray}{73.37 ± 23.77 }   & \textcolor{gray}{78.52 ± 11.84}    & \textcolor{gray}{25.46 ± 13.99} & \textcolor{gray}{43.61 ± 52.26} \\
% wICP            & -- ± --    & -- ± --    & -- ± -- & -- ± -- \\
% GeoTransformer  & 103.07 ± 38.50    & 115.05 ± 24.16    & 109.01 ± 18.18 & 92.04 ± 16.83 \\
% RegTr           & -- ± --    & -- ± --    & -- ± -- & -- ± -- \\
GCNNet          & 23.16 ± 13.30    & 26.03 ± 11.67    & \textcolor{gray}{32.08 ± 10.10 }& \textcolor{gray}{31.20 ± 10.83} \\
LiverMatch      &13.94 ± 13.08    & 14.57 ± 12.43    & 12.33 ± 3.95 & 13.45 ± 2.59 \\

% Nonrigid Registration Group
% \addlinespace[0.5em]
% \multicolumn{4}{@{}l}{\textbf{Nonrigid Registration}} \\
\cmidrule(){1-5}
CPD            & \textcolor{gray}{48.68 ± 16.50}    & \textcolor{gray}{46.74 ± 10.84}    & \textcolor{gray}{36.41 ± 4.61} & \textcolor{gray}{41.14 ± 6.45}  \\
HMM-CPD        & 25.53 ± 15.70    &26.31 ± 15.59    & \textcolor{gray}{29.44 ± 3.09} & \textcolor{gray}{39.40 ± 9.43} \\
HMM-CPD (aniso)  & 23.47 ± 15.27    & 26.08 ± 18.68    & \textcolor{gray}{28.02 ± 3.02} & 25.29 ± 5.23 \\
% BCPD            & 72.60 ± 30.71    & 76.25 ± 20.73    & 53.02 ± 3.65 & 48.07 ± 6.04  \\
% Deep-Geo-Reg    & 91.8 ± 46.5    & -- ± --    & -- ± -- & -- ± --  \\
% GraphSCNet         & -- ± --    & -- ± --    & -- ± -- & -- ± --  \\

Robust-DefReg   & 26.40 ± 16.10    & 29.07 ± 15.44    & \textcolor{gray}{35.02 ± 1.45} & \textcolor{gray}{42.87 ± 3.20} \\
LiverMatch + NDP  & 19.60 ± 18.08    & 16.60 ± 12.13    & \textcolor{gray}{20.12 ± 2.92} & 22.34 ± 3.53  \\
Lepard         & 19.92 ± 14.73    & 21.81 ± 14.07    & 16.34 ± 7.35 & 19.28 ± 6.00 \\
C2P-Net         & 19.15 ± 13.73    & 19.17 ± 12.57    & 17.00 ± 1.07 & 22.21 ± 4.22  \\
PBSM            & 27.59 ± 29.02    & 30.28 ± 26.81    & 10.03 ± 6.02 & 8.59 ± 5.39  \\
BCF\_FEM         & 21.90 ± 30.47    & 18.11 ± 22.49    & \underline{9.02 ± 6.34} & \underline{6.72 ± 2.88} \\
V2S-Net         & \underline{6.36 ± 6.48}    & \underline{6.25 ± 6.04}    & 9.03 ± 4.06 & 7.21 ± 1.86  \\
PIVOTS (ours) & \textbf{3.11} ± \textbf{3.38}    & \textbf{4.06} ± \textbf{3.35}    & \textbf{5.96} ± \textbf{1.47} & \textbf{6.38} ± \textbf{2.56}  \\
\bottomrule
% \label{exp:table_mean_registration_error}
\end{tabularx}

% \begin{tablenotes}
% \small
% \item Note: All experiments were conducted on an NVIDIA A100 GPU with 40GB memory. ICP: Iterative Closest Point, FFD: Free-Form Deformation, LDDMM: Large Deformation Diffeomorphic Metric Mapping. Reported values represent mean ± standard deviation over 5 independent runs.
% \end{tablenotes}
\end{table*}

% BCPD: https://github.com/ohirose/bcpd?tab=readme-ov-file#macos-and-linux
% Robust-DefReg: https://github.com/m-kinz/Robust-DefReg/blob/main/Robust-DefReg.ipynb
% DDFR: https://gitlab.kuleuven.be/u0132345/deepdiffeomorphicfaceregistration

\subsection{Comparison with State-of-the-Art Methods}
\label{exp:comparison}
We select widely adopted registration methods from the general computer vision and liver registration communities as our baseline methods. They can be roughly categorized into rigid methods and non-rigid methods.  ICP~\citep{besl1992icp} serves as a rigid, non-learning baseline. GCNNet~\citep{zhu2022neighborhood} and LiverMatch~\citep{yang2023learning} are rigid learning-based approaches. Classical non-rigid, non-learning methods include CPD~\citep{myronenko2010point} and HMM-CPD~\citep{min2019generalized, min2020feature}, while PBSM~\citep{suwelack2014physics} and BCF\_FEM~\citep{Yang2024BoundaryCB} represent biomechanics-based non-rigid methods. Learning-based non-rigid methods include Lepard~\citep{li2022lepard}, Robust-DefReg~\citep{monji2024robust}, LiverMatch+NDP~\citep{li2022non}, C2P-Net~\citep{liu2024non}, and PIVOTS.
% , including ICP, GCNNet, and LiverMatch, and non-rigid methods, consisting of CPD, HMM-CPD, Lepard, Robust-DefReg, LiverMathc+NDP, C2P-Net, PBSM, BCF\_FEM and V2S-Net.  further divided into traditional and learning-based methods. 
The hyperparameters of all methods are carefully optimized, and all learning-based methods are trained using the same synthetic training dataset as the proposed method. For PBSM and BCF\_FEM, the preoperative surfaces must be meshed with volume elements, which we achieve with the GMSH~\citep{gmsh_2009} software. Because the speed of these methods strongly depends on the number of elements in the meshes, we downsample surfaces to a maximum of roughly 5000 points.

% TODO: Incorporate something like this when describing PBSM results:
% PBSM: "PBSM results are sensitive to the \emph{charge} hyperparameter, which may have to be chosen according to the current deformation (PRD) and partial visible surface. Since this is infeasible for our large dataset, we instead apply PBSM separately for the charges 100, 500, 1000 and 2000 to each sample, calculate the TRE for each result and report only this value. While this may give a slightly unfair advantage to PBSM, this approach replaces the need to manually fine-tune for every sample."

The results of all methods are listed in Tab.~\ref{tab:registration_performance}, which presents a quantitative comparison of registration performance among a diverse set of baseline methods across four datasets: Synthetic, AMOS, Phantom, and HHLBM. %The evaluation metric is MED and TRE, reported as mean ± standard deviation in millimeters, computed based on ground truth displacements (Synthetic and AMOS) or sparse landmark (Phantom and HHLBM).
As evaluation metric we use MED computed based on ground truth displacements (Synthetic and AMOS) and TRE calculated from sparse landmarks (Phantom and HHLBM), reported as mean ± standard deviation in millimeters. 
For reference, the pre-registration displacement (PRD) is included to indicate the initial misalignment magnitude in each dataset prior to registration.

Overall, the proposed method achieves the lowest mean registration error and Mean + SD across all datasets, outperforming both rigid and non-rigid, traditional and learning-based approaches. Notably, the performance gain is most pronounced in the Synthetic and AMOS datasets, where the average error ($3.11$\,mm and $4.06$\,mm) is reduced by a substantial margin compared to the next-best methods. In particular, besides LiverMatch, V2S-Net, and our method, the other methods obtain much worse registration errors %, which are 
around $20$\,mm and much higher, although learning-based methods are trained on the same dataset and iterative methods are fine-tuned as much as possible. For example, ICP ($73.37$\,mm and $78.52$\,mm), CPD ($48.68$\,mm and $46.74$\,mm), and Robust-DefReg ($26.40$\,mm and $29.07$\,mm) %are typical failure methods
appear to struggle consistently. These results indicate that volume-to-surface registration is a challenging task and requires dedicated designs to %tackle the internal structure and high inconsistency between the source and target.
tackle the high inconsistency between source and target and accurately infer the movement of internal structures.% distant from the available partial surface information. %Hence, these results and comparisons highlight the effectiveness of our method in registering diverse volume shapes of the organs. The RMSE of PIVOTS on the Synthetic and AMOS datasets is $4.32 ± 4.35$ and $4.96 ± 4.13$ for reference.
The comparison between the Synthetic and AMOS datasets also illustrates PIVOTS' capability to generalize from random shapes during training to real liver shapes at inference time, with an increase in MED from $3.11 \text{±} 3.38$\,mm to $4.06 \text{±} 3.35$\,mm and RMSE from $4.32 \text{±} 4.35$\,mm to $4.96 \text{±} 4.13$\,mm, respectively.

Furthermore, on datasets with real deformation, namely the Phantom and HHLBM datasets, our method similarly outperforms others with a registration error of $5.96$\,mm and $6.38$\,mm, respectively, suggesting higher robustness and ability of the method in dealing with real noise and deformation compared to baseline methods, e.g., V2S-Net ($9.03$\,mm and $7.21$\,mm). Beyond absolute performance, the consistency and stability of each method across datasets with various noise and deformation is also critical and significant in practice to ensure safe navigation even when conditions in clinical application may differ. For instance, some methods such as Robust-DefReg and NDP show substantial variation in performance depending on the dataset, suggesting sensitivity to specific deformation characteristics or anatomical context. In contrast, our method demonstrates a stable and low error across all domains, reflecting effective generalization from synthetic to real anatomical deformation and across different evaluation protocols (dense field vs. sparse landmarks).

In summary, several insights can be obtained:

\textbf{Volume-to-Surface Registration}: For datasets with real deformation, only volume-to-surface baseline methods achieve a registration error lower than $1$\,cm, e.g. PBSM and V2S-Net, indicating the criticality of volume-to-surface methods in order to obtain high accuracy of the prediction for internal structures. Furthermore, deep learning methods originally designed for surface-to-surface registration, e.g. Lepard and C2P-Net, require dedicated modification to adapt to such a different task in liver navigation. 

\textbf{Necessity of Non-rigid Registration}: Although LiverMatch achieves %TREs
registration errors lower than the majority of non-rigid methods across all four datasets, none of the mean TREs undercut %breakthrough 
the $1$\,cm threshold, which is only achieved by four of the nonrigid methods. %Also, the much worse results from ICP and GCN-Net prove the significance of non-rigid registration.

\textbf{Synthetic and Real Domain Gap}: Our method obtained a mean registration error lower than $4$\,mm on datasets with synthetic deformation. However, the performance degrades on the other two real datasets by around $2$\,mm. This degradation hints at %presents 
a noticeable gap between synthetic and real datasets, %including different material models and noise. 
which could be attributed to different materials or properties of the added noise. An alternative influence could be the nature of the deformation: In the synthetic data, the organ can be stretched or folded. The main deformation in the Phantom dataset stems from a US probe causing a local perturbation and the in vivo deformation of the HHLBM dataset is limited by the undisturbed relation of the liver to the surrounding organs and tissues. Especially in later stages of a surgery, one would expect higher deformations than in both of those cases. Therefore, while the performance differences can inform room for improvement in the training data generation, an intraoperative dataset would be highly valuable in assessing the generalization ability to true intraoperative deformations in more detail.

\textbf{Inference Speed}: The methods based on biomechanical simulation (PBSM and BCF\_FEM) obtain very good registration results on some samples, especially when hyperparameters are fine-tuned and the initial displacement (PRD) is low. However, they are orders of magnitudes slower and often take more than half a minute for a single sample, while our PIVOTS method roughly takes 230 ms.
% \textcolor{red}{200 ms TODO}.

\subsection{Deformation Level Experiment}
\label{exp:deformation_levels}
% \begin{figure}[!ht]
%     \centering
%     \includegraphics[trim=0cm 0cm 0cm 0cm,clip,width=0.5\linewidth]{pics/TRE_vs_PRD_multiple.pdf}
%     \caption{Placeholder for deformation experiment.}
%     \label{fig:deform_level}
% \end{figure}

% \begin{figure}[ht]
%     \centering
%     \subfloat[First image caption]{%
%         \includegraphics[width=0.45\linewidth]{pics/TRE_vs_PRD_multiple_syn.pdf}%
%         \label{fig:subfig1}
%     }
%     \hfill
%     \subfloat[Second image caption]{%
%         \includegraphics[width=0.45\linewidth]{pics/TRE_vs_PRD_multiple_amos.pdf}%
%         \label{fig:subfig2}
%     }
%     \caption{Deformation level experiments, left generated datasets, right amos dataset}
%     \label{fig:deform_level}
% \end{figure}

\begin{figure}[!ht]
    \centering
    \includegraphics[trim=0cm 0cm 0cm 0cm,clip,width=\linewidth]{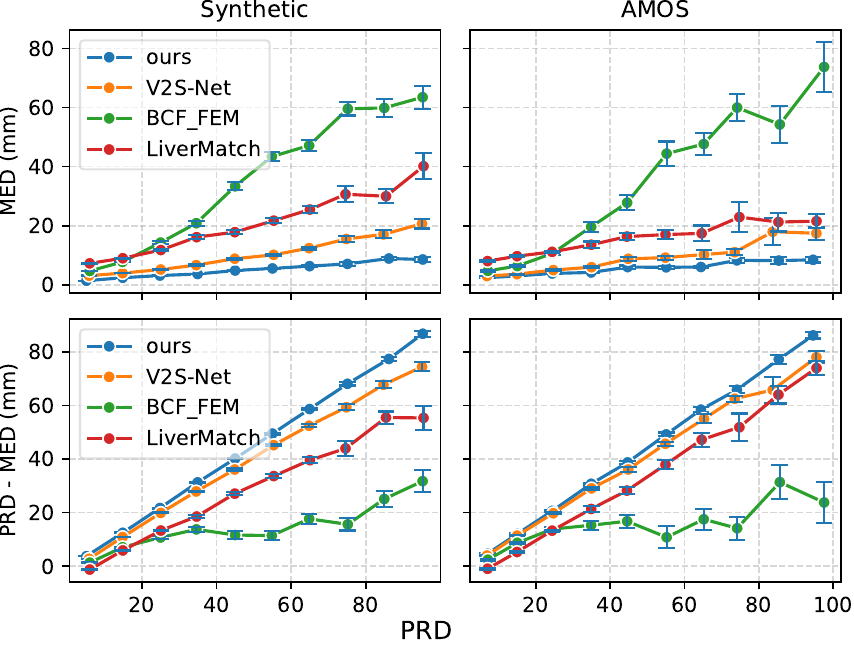}
    \caption{Comparison of methods on various deformation levels (PRD) on the synthetic (left) and AMOS (right) datasets. Top row: Mean Euclidean distance (MED) versus PRD, demonstrating that our method exhibits the greatest robustness to initial misalignment. Bottom row: relative registration error (MED - PRD) versus PRD, showing that our approach achieves the most pronounced error reduction.}
    \label{fig:deform_level_syn_amos}
\end{figure}

\begin{table}[ht]
\centering
\caption{Pearson and Spearson metrics for Synthetic and AMOS datasets}
\label{tab:deform_level_pearson}
\small % or \scriptsize if needed to fit
\begin{tabular}{lcc|cc}
\toprule
\textbf{Method} 
& \multicolumn{2}{c|}{\textbf{Synthetic}} 
& \multicolumn{2}{c}{\textbf{AMOS}} \\
\cmidrule(lr){2-3} \cmidrule(lr){4-5}
& Pearson & Spearman 
& Pearson & Spearman \\
\midrule
BCE\_FEM    & 0.437 & 0.557 & 0.359 & 0.561 \\
LiverMatch  & 0.864 & 0.904 & 0.924 & 0.919 \\
V2S-Net     & 0.982 & 0.985 & 0.981 & 0.985 \\
ours        & 0.996 & 0.995 & 0.993 & 0.991 \\
\bottomrule
\end{tabular}
\end{table}

% During laparoscopic surgery, the liver undergoes large deformation caused by respiration, tool manipulation, and pneumoperitoneum. It is critical to recover the intraoperative liver status under large deformation. To quantify the registration performance of different methods under different deformation levels using the generated and AMOS test sets, we visualize the changes of registration errors over the deformation of the samples using the generated and AMOS test sets in Fig.~\ref{fig:deform_level_syn_amos} and the Pearson correlation coefficients in Tab.~\ref{tab:deform_level_pearson}.

Since the liver undergoes large deformation during surgery, e.g., due to tool manipulation, it is critical for the registration to reliably capture little as well as highly deformed states. We use the synthetic and AMOS test sets to quantify the influence of different deformation levels on the registration performance of representative methods from each group in Tab.~\ref{tab:registration_performance} as visualized in Fig.~\ref{fig:deform_level_syn_amos}, Also, we present the Pearson correlation coefficients in Tab.~\ref{tab:deform_level_pearson}.

From Fig.~\ref{fig:deform_level_syn_amos}, mean registration errors of our methods increase steadily with a relatively low slope, ending with errors below $10$\,mm for the largest deformation group ($>80$\,mm). In comparison, the other baseline methods present a more inclined trend with the increase of the deformation level, especially BCF\_FEM, the mean error reaching more than $20$\,mm for a deformation larger than $40$\,mm. As a rigid registration method, LiverMatch surpasses BCF\_FEM on both test sets and achieves similar errors as V2S-Net when presented with large deformations in the AMOS test set, with an MED of $20$\,mm.

The relative registration error is computed as $\Delta_{RE} = PRD - MED$, showing how much the registration error increases or decreases compared to the initial error prior to registration. Tab.~\ref{tab:deform_level_pearson} exhibits different dependencies of relative registration errors on initial error using Pearson correlation coefficients $r$ and Spearman correlation coefficients $\rho$.
Our method achieves near-perfect linear ($r= 0.996$) and rank-order ($\rho = 0.995$) correlations, indicating that larger initial misalignments are almost corrected by proportionally larger amounts. 
V2S-Net follows closely ($r = 0.981$, $\rho = 0.9850$), likewise demonstrating a very tight coupling between case difficulty and error reduction. In contrast, BCF\_FEM shows a substantially weaker relationship ($r = 0.437$, $\rho = 0.557$), reflecting a more equilibrated strategy that yields more uniform improvements regardless of initial error magnitude. As a rigid alignment method, LiverMatch occupies an intermediate position ($r = 0.8643$, $\rho = 0.9043$), balancing targeted correction of challenging cases with non-negligible gains for easier ones. 
These results suggest that our method and V2S-Net yield proportional registration error correction, BCF\_FEM presents global consistency regardless of the initial misalignment, and LiverMatch shows a comparable capability in error reduction.
Overall, our methods exhibit superior performance regarding dealing with both small and large deformations.

\subsection{Noise Level Experiments}
\label{exp:noise_levels}

\begin{figure*}[ht!]

\centering

\renewcommand{\arraystretch}{1.5}
\arrayrulecolor{lightgray}
\begin{tabular}{>{\centering\arraybackslash}m{0.5cm} 
                >{\centering\arraybackslash}m{3.5cm} 
                >{\centering\arraybackslash}m{3.5cm} 
                >{\centering\arraybackslash}m{3.5cm} 
                >{\centering\arraybackslash}m{3.5cm}}
%\hline
 & \text{$A_P = 0$\,mm} & \text{$A_P = 3$\,mm} & \text{$A_P = 9$\,mm} & \text{$A_P = 15$\,mm} \\
%\hline
%\rotatebox{90}{\makebox[height][c]{Your Text}}
\rotatebox[origin=c]{90}{\makebox[\height][c]{$\sigma = 0$\,mm}} &
% \frame{
\includegraphics[width=3.5cm,trim=30mm 50mm 0mm 20mm,clip]{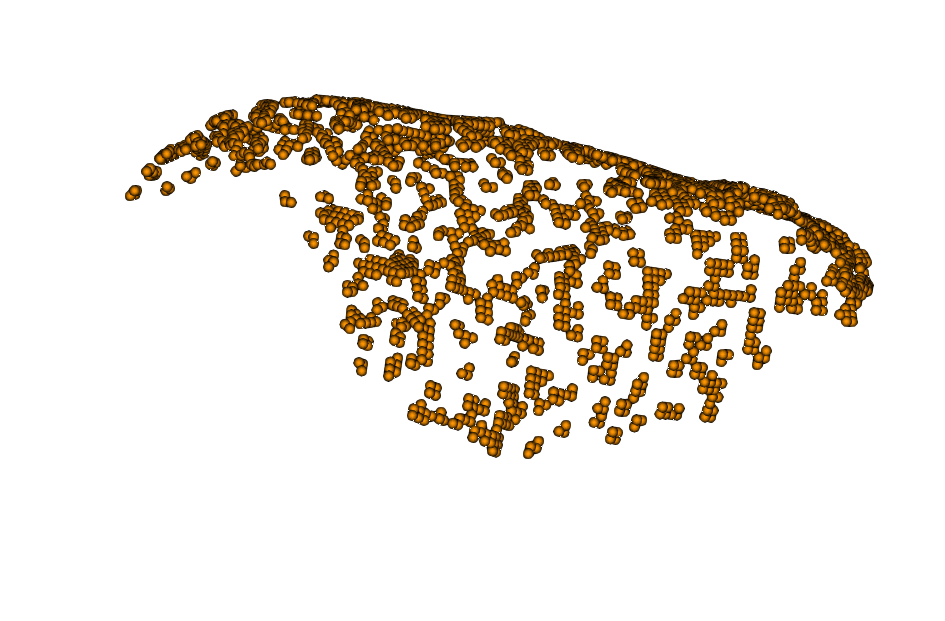} & 
\includegraphics[width=3.5cm,trim=30mm 50mm 0mm 20mm,clip]{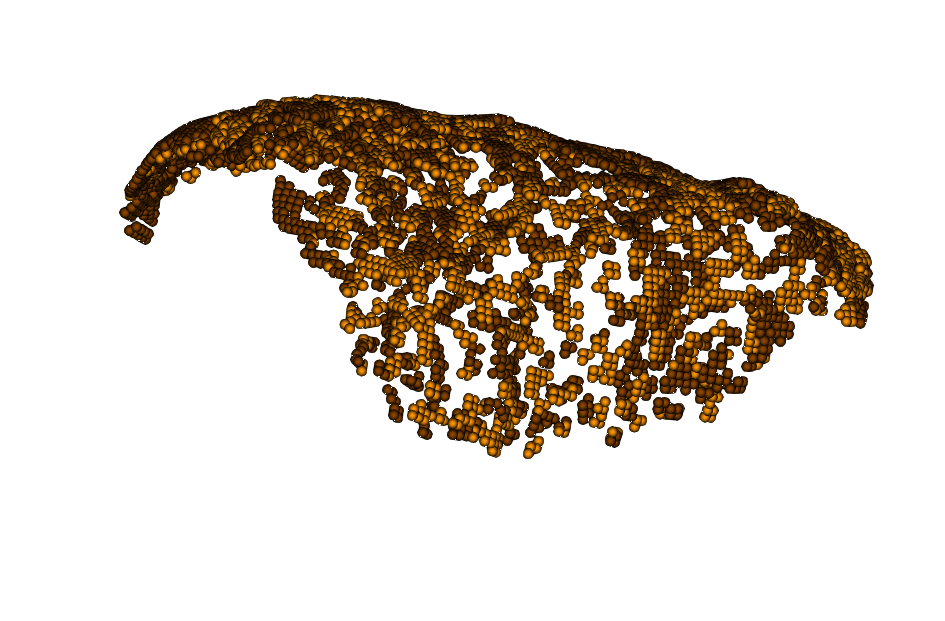} & 
\includegraphics[width=3.5cm,trim=25mm 50mm 0mm 20mm,clip]{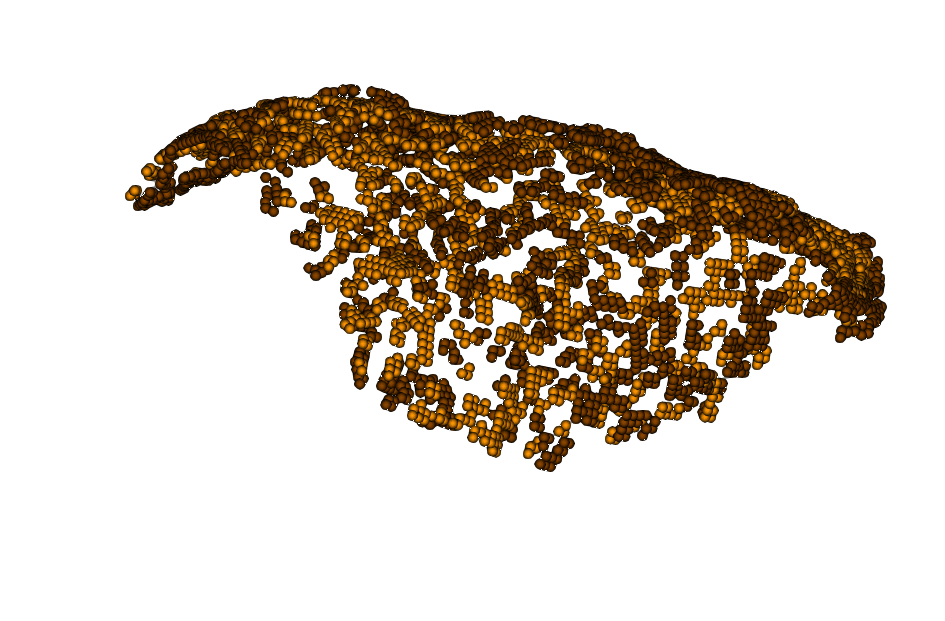} & 
\includegraphics[width=3.5cm,trim=30mm 50mm 0mm 20mm,clip]{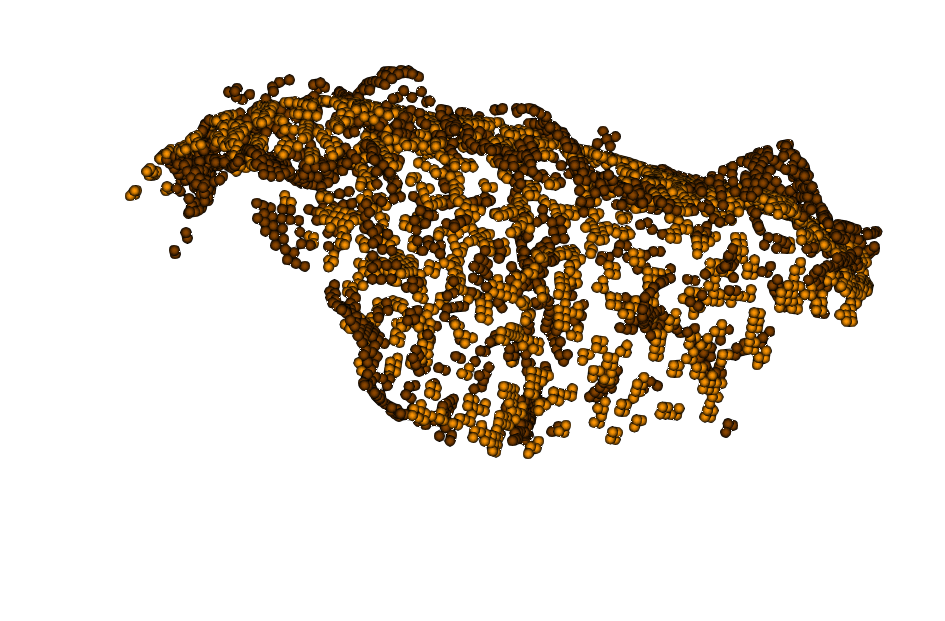} \\
% \hline
\rotatebox[origin=c]{90}{\makebox[\height][c]{$\sigma = 2.5$\,mm}} & 
\includegraphics[width=3.5cm,trim=30mm 39mm 0mm 20mm,clip]{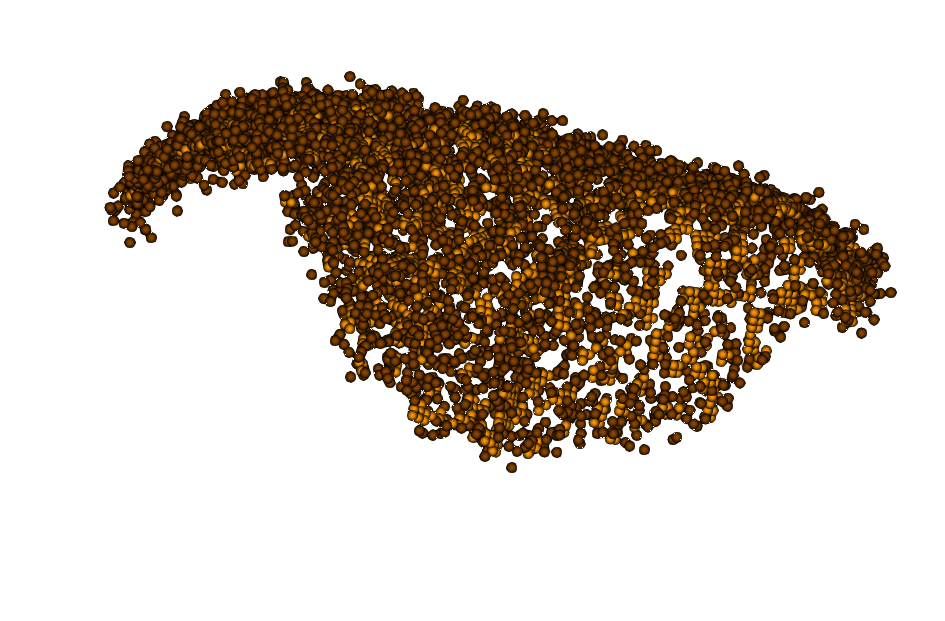} & 
\includegraphics[width=3.5cm,trim=30mm 39mm 0mm 20mm,clip]{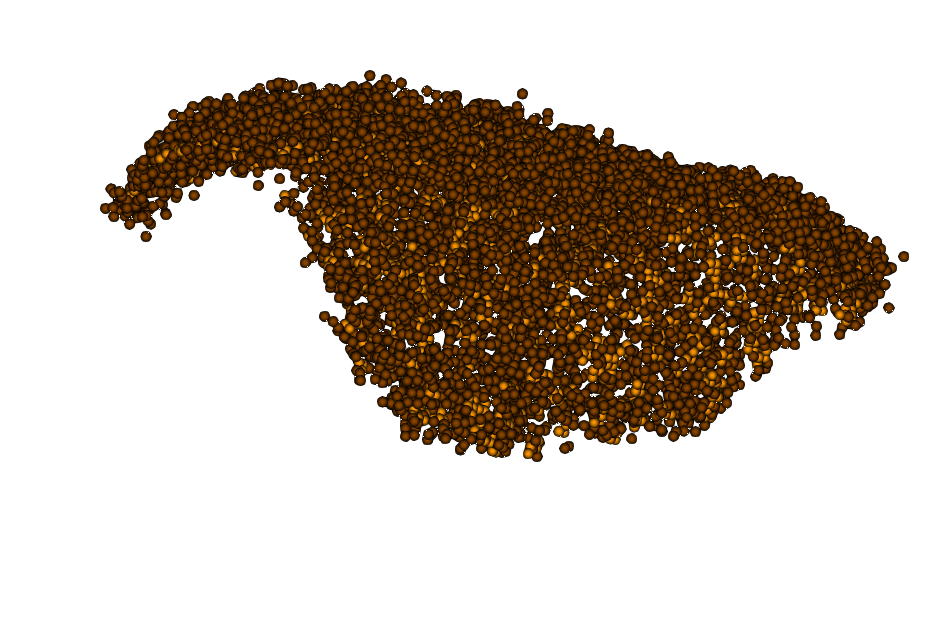} & 
\includegraphics[width=3.5cm,trim=25mm 39mm 0mm 20mm,clip]{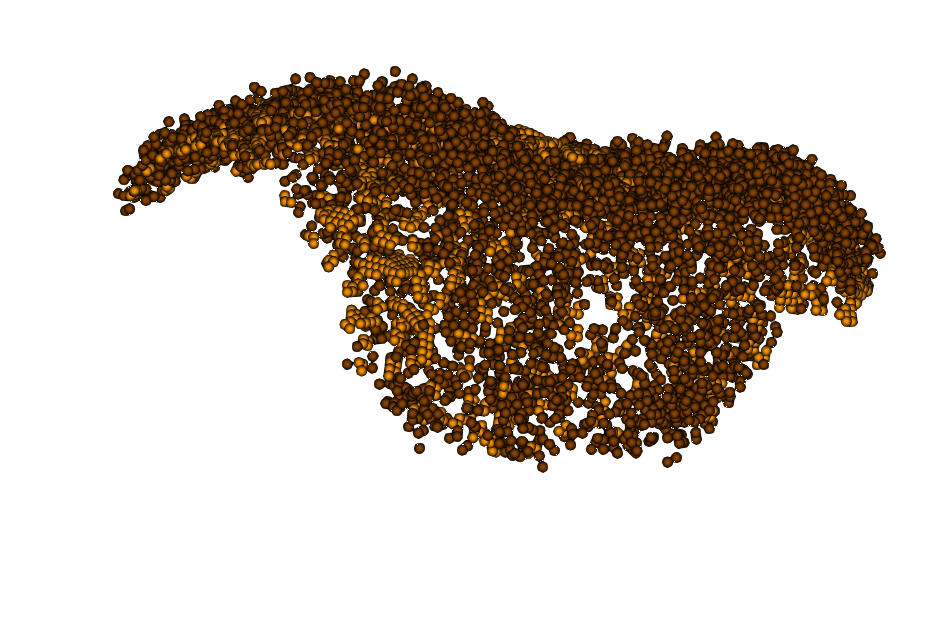} & 
\includegraphics[width=3.5cm,trim=30mm 39mm 0mm 20mm,clip]{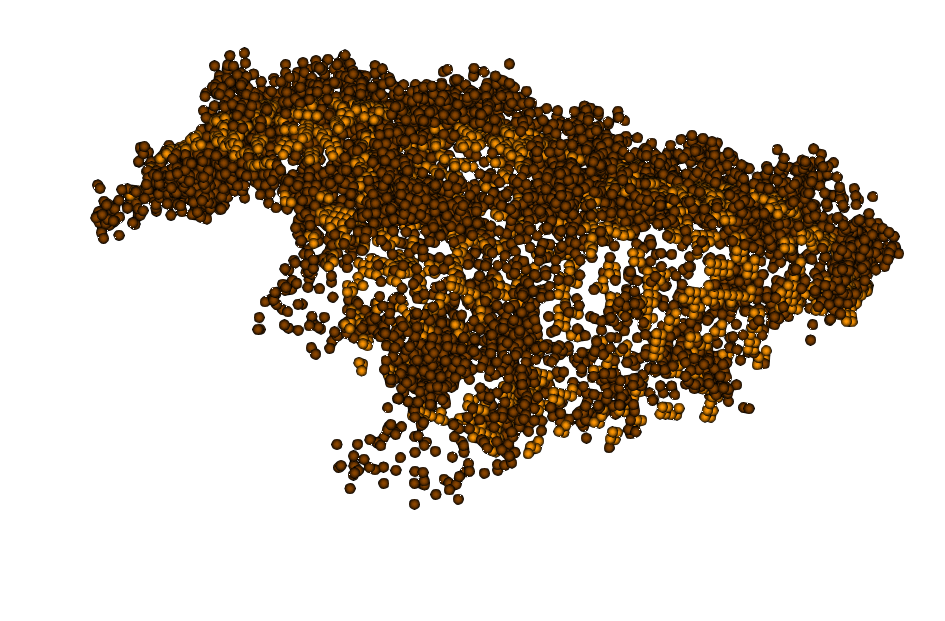} \\
%\hline
\rotatebox[origin=c]{90}{\makebox[\height][c]{$\sigma = 5$\,mm}} & 
\includegraphics[width=3.5cm,trim=30mm 40mm 0mm 20mm,clip]{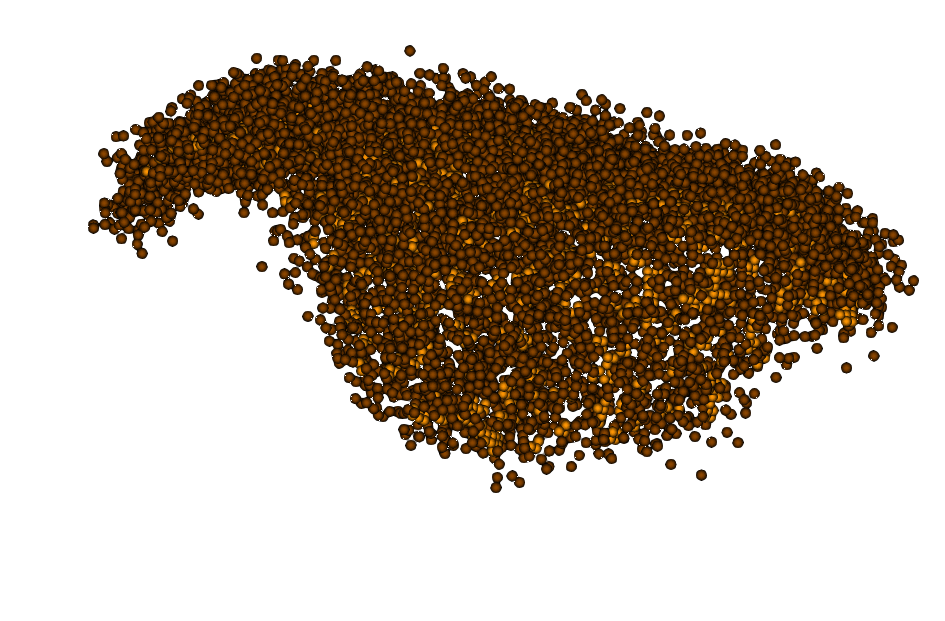} & 
\includegraphics[width=3.5cm,trim=30mm 40mm 0mm 20mm,clip]{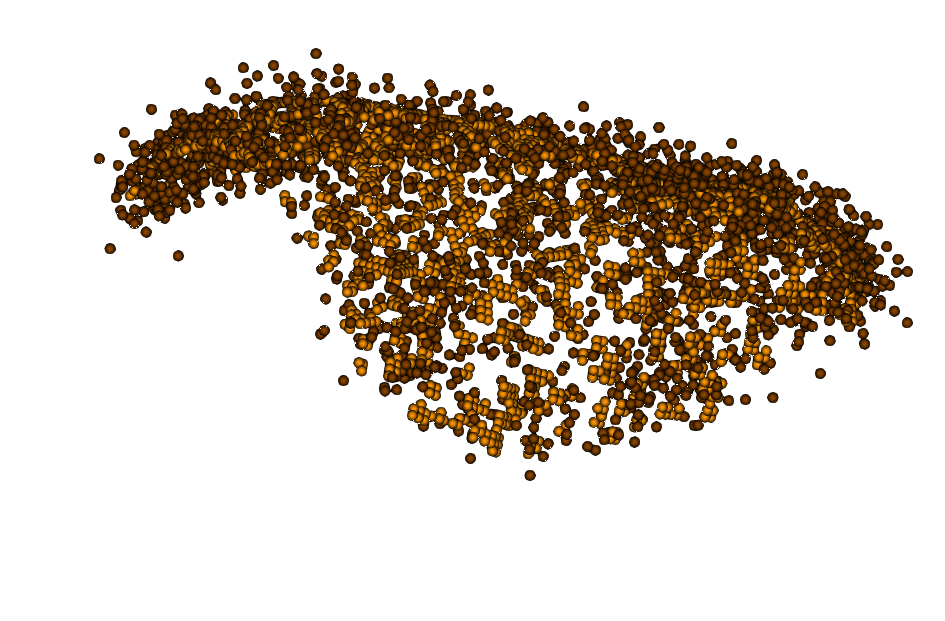} & 
\includegraphics[width=3.5cm,trim=25mm 40mm 0mm 20mm,clip]{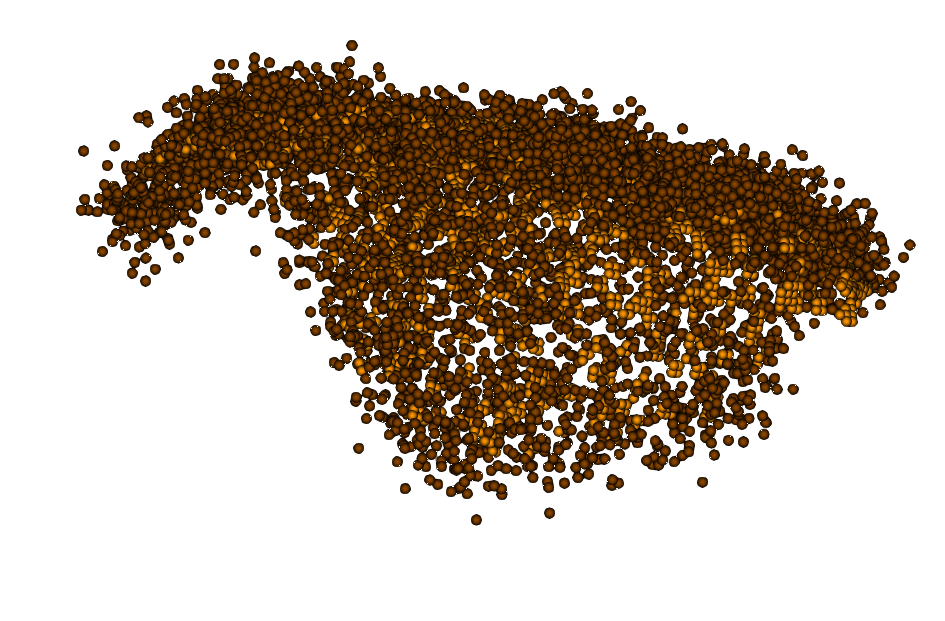} & 
\includegraphics[width=3.5cm,trim=30mm 40mm 0mm 20mm,clip]{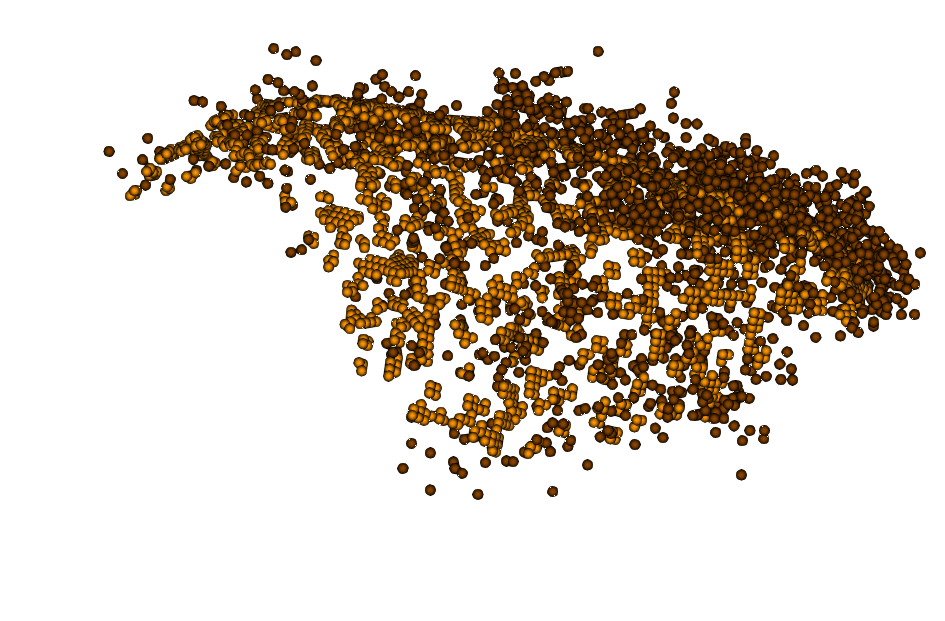} \\
%\hline
\end{tabular}
\caption{Input for the noise level experiment. Example partial intraoperative surface $\mathbf{S}$ from the AMOS$_\text{noise}$ dataset (orange) with different combinations of Gaussian noise with deviation $\sigma$ (rows) and Perlin noise with amplitude $A_P$ (columns). Noise levels (brown) increase from left to right and from top to bottom. Gaussian noise blurs the intraoperative point cloud while Perlin noise causes nonuniform local distortions to the apparent surface shape. The randomized subsampling of the point cloud after adding noise additionally increases the difficulty.}
\label{fig:noise_levels}
\end{figure*}
\arrayrulecolor{black}

\begin{figure*}[!ht]
    \centering
    \includegraphics[trim=0cm 0cm 0cm 0cm,clip,width=\textwidth]{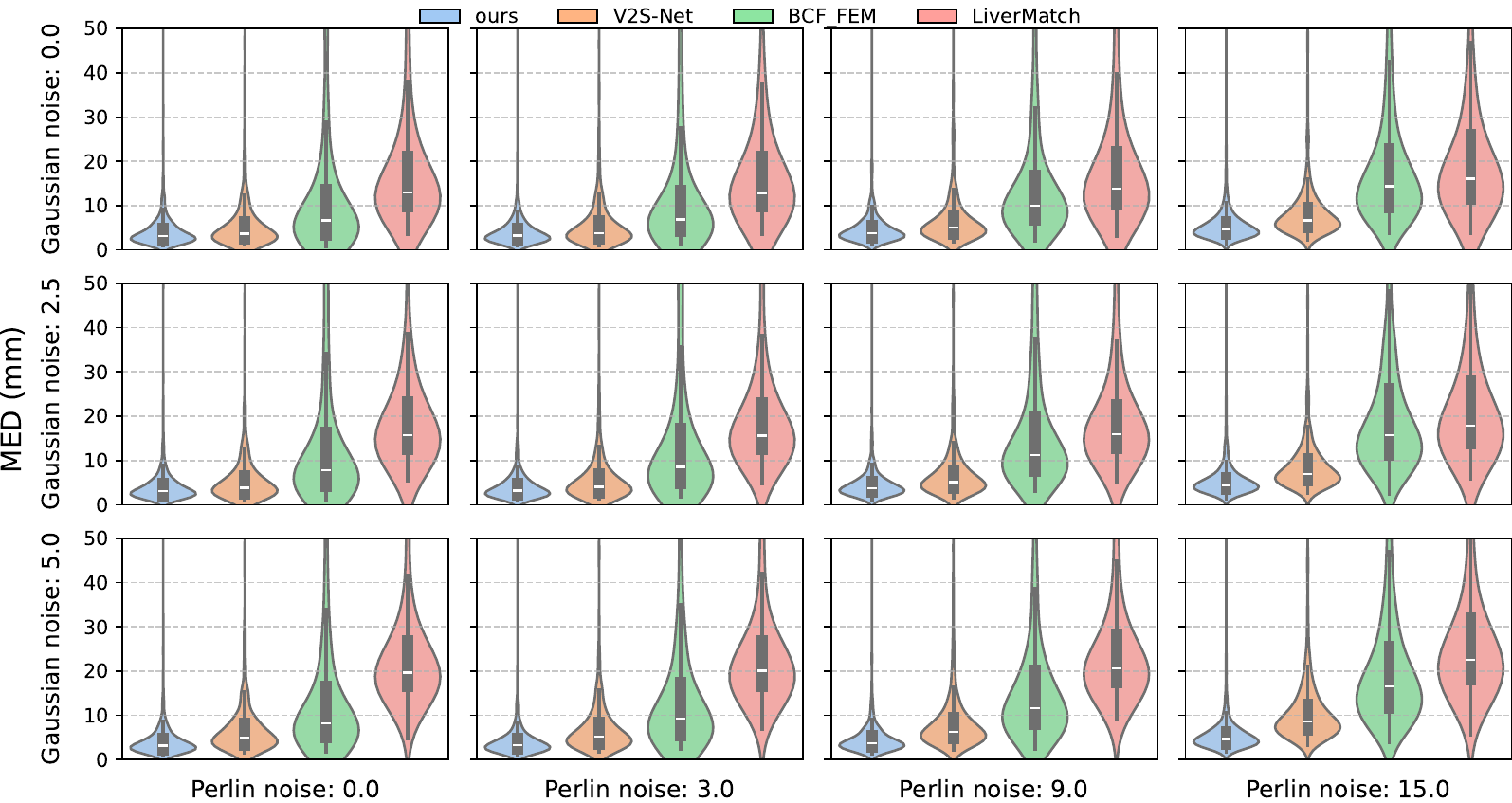}
    \caption{Registration error distributions on various noise levels. Two types of noise are involved: Perlin noise (horizontal, range: $A_P \in \{0, 3, 6, 9, 12, 15\}\text{ mm}$) and Gaussian noise (vertical, range: $\sigma \in \{0, 2.5, 5.0\} \text{ mm}$). Each subplot exhibits the registration error distribution of four methods. Our method presents the most robustness against various amounts and types of noise. }
    \label{fig:amos_noise_level}
\end{figure*}

% \begin{figure}[!ht]
%     \centering
%     % \frame{
%     \includegraphics[trim=0cm 7.4cm 12.1cm 0cm,clip,width=\linewidth]{pics/amos_noise_level_qualitative.pdf}
%     % }
%     \caption{Qualitative comparison between results of PIVOTS on intraoperative surfaces with various Gaussian noise (first row) and Perlin noise amplitudes (second row). The input and ground truth deformation are demonstrated on the left. % in the first column. 
%     PIVOTS exhibits strong robustness against various amounts of noise.}
%     \label{fig:amos_noise_levels_qualitative}
% \end{figure}

\begin{figure}[!ht]
    \centering
    % \frame{
    \includegraphics[trim=0cm 1cm 1.7cm 0cm,clip,width=\linewidth]{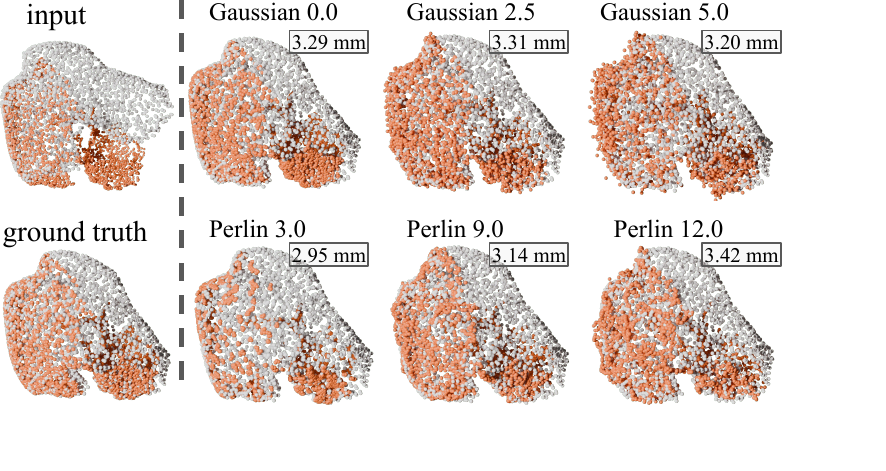}
    % }
    \caption{Qualitative comparison between results of PIVOTS on intraoperative surfaces with various Gaussian noise (first row) and Perlin noise amplitudes (second row). The input and ground truth deformation are demonstrated on the left. % in the first column. 
    PIVOTS exhibits strong robustness against various amounts of noise.}
    \label{fig:amos_noise_levels_qualitative}
\end{figure}

In 3D-3D liver registration scenarios, the reconstructed liver surfaces using SfM or SLAM from the laparoscopic videos may be corrupted due to various noise or artifacts, substantially degrading the registration accuracy. To quantify the impact of various noise levels, we generated 18 noisy partial surfaces from the AMOS dataset by superimposing Perlin noise (amplitudes $A_P \in \{0, 3, 6, 9, 12, 15\}\text{ mm}$) and Gaussian noise ($\sigma \in \{0, 2.5, 5.0\} \text{ mm}$) as illustrated in figure \ref{fig:noise_levels}. 

We selected three representative methods with relatively low registration error in Tab.~\ref{tab:registration_performance} for this experiment, including V2S-Net, BCF\_FEM and LiverMatch. 

Fig.~\ref{fig:amos_noise_level} illustrates all registration results, organized by noise combinations, where the horizontal axis represents Perlin noise and the vertical direction corresponds to Gaussian noise. Overall, PIVOTS maintains a consistently low registration error across all noise combinations, showing strong robustness against various types and amounts of noise. On the other hand, V2S-Net and BCF\_FEM present noticeable and rapid degradation with more noise, although the baseline methods work reasonably well with low amounts of noise. In contrast, the errors of PIVOTS show a much steadier deteriorating trend. In particular, the impact of Perlin noise is remarkably larger than Gaussian noise. This finding aligns well with the visualization in Fig.~\ref{fig:amos_noise_level}, where Gaussian noise makes the partial surface more fuzzy while the basic geometry remains, while on the other hand, target partial surfaces present a bumpy appearance under higher Perlin noise, potentially modifying the geometry and complicating the correspondence exploration for registration methods. 
These trends are also visible in the qualitative results, illustrated for a subset of the noise levels in Fig.~\ref{fig:amos_noise_levels_qualitative}: PIVOTS yields stable results with increasing amounts of Gaussian noise and slightly rising errors with higher Perlin noise amplitudes, but the results are qualitatively promising.

Besides the mean error, the development of the error variability further underpins PIVOTS's strong noise handling capabilities. The top left plot (without noise) shows V2S-Net and PIVOTS clearly outperforming BCF-FEM and LiverMatch while on a similar level to each other. Despite BCF-FEM achieving a mean MED below $1$\,cm, the error distribution shows roughly a third of samples with an error above this value, making the method unreliable for a considerable subset of patients even in the ideal noise-free setting. With increasing noise levels, the error distribution spreads for all methods in general. However, PIVOTS still manages to contain the error for all but a small fraction of its predictions below the $1$\,cm threshold even at the highest noise level, giving it a strong edge over the V2S-Net baseline as well.  %Taking a group of noise as a qualitative example in 
%Illustrated for a subset of the noise levels in Fig.~\ref{fig:amos_noise_levels_qualitative}, PIVOTS yields stable results with increasing amounts of Gaussian noise, and slightly rising errors with higher Perlin noise amplitudes, but the results are qualitatively promising. These trends align well with the distributions in Fig.~\ref{fig:amos_noise_level}.

Additionally, this experiment allows the conclusion that PIVOTS is able to generalize between different surface extraction methods, as it was trained only on randomly extracted surfaces but AMOS$_{\text{noise}}$ contains camera-like partial surfaces, and the MED of this experiment remains comparable to the second column of table \ref{tab:registration_performance}.

\subsection{Intraoperative Visibility Experiments}
\label{exp:visiblity}

% \begin{figure}[!ht]
%     \centering
%     \includegraphics[trim=0cm 0cm 0cm 0cm,clip,width=\linewidth]{pics/TRE_vs_visibility_comparison.pdf}
%     \caption{Placeholder for visibility experiment.}
%     \label{fig:visibility_nki_hhlbm}
% \end{figure}

% \begin{table}[ht]
% \centering
% \scriptsize
% \begin{tabularx}{\columnwidth}{l*{6}{>{\centering\arraybackslash}X}}
% \toprule
% Dataset 
%   & \multicolumn{3}{c}{Phantom}
%   & \multicolumn{3}{c}{HHLBM} \\
% \cmidrule(lr){2-4} \cmidrule(lr){5-7}
% vis (\%)
%   & \mbox{0~--~10}
%   & \mbox{10~--~20}
%   & \mbox{20~--~30}
%   & \mbox{0~--~10}
%   & \mbox{10~--~20}
%   & \mbox{20~--~30} \\
% \midrule
% PBSM    & \mbox{22.30±6.83}   & 14.84 6.93    & 9.87 5.17     & 10.18 5.77     & 0            & 0 \\
% BCF\_FEM & 10.22 6.22  & 0 9.32 6.06   & 8.89  5.61     & 9.49 6.57     & 0              & 0 \\
% V2S-Net & 13.73  4.88   & 0 9.88 4.43    & 8.91 4.18     & 9.36 3.99     & 0              & 0 \\
% ours    & 9.33 ± 7.29     & 6.71 2.68    & 6.46 1.47    & 6.35 127     & 0             & 0 \\
% \bottomrule
% \end{tabularx}
% \caption{Placeholder for visibility experiment}
% \label{tab:visibility_tre}
% \end{table}

\begin{table}[ht]
\centering
\caption{TREs (mm) of visibility experiment on the Phantom and HHLBM datasets}
\scriptsize
\begin{tabularx}{\columnwidth}{%
    >{\raggedright\arraybackslash}X   % Dataset
    >{\centering\arraybackslash}X     % Visibility
    *{4}{>{\centering\arraybackslash}X}% Methods
}
\toprule
Dataset        & \mbox{Vis. Per. (\%)}      & PBSM & BCF\_FEM & V2S-Net & ours \\
\midrule
\multirow{4}{*}{Phantom}
               & \mbox{0~--~10}       & \mbox{22.30 ± 6.83}    & \mbox{10.22 ± 6.62}    & \mbox{13.73 ± 4.88}  & \textbf{\mbox{9.33 ± 4.23}}  \\
               & \mbox{10~--~20}      & \mbox{14.84 ± 6.93}    & \mbox{9.32 ± 6.06}     & \mbox{9.88 ± 4.43}   & \textbf{\mbox{6.71 ± 2.68}} \\
               & \mbox{20~--~30}      & \mbox{9.87 ± 5.17}     & \mbox{8.89 ± 5.61}     & \mbox{8.91 ± 4.18}   & \textbf{\mbox{6.46 ± 1.47}} \\
               & \mbox{30~--~40}      & \mbox{10.18 ± 5.77}    & \mbox{9.49 ± 6.57}     & \mbox{9.36 ± 3.99}   & \textbf{\mbox{6.35 ± 1.27}} \\
\midrule
\multirow{4}{*}{HHLBM}
               & \mbox{0~--~10}       & \mbox{20.56 ± 6.28}    & \mbox{18.61 ± 5.67}     & \textbf{\mbox{18.09 ± 7.29}}   & \mbox{19.93 ± 9.92} \\
               & \mbox{10~--~20}      & \mbox{16.04 ± 5.78}    & \mbox{13.03 ± 6.43}     & \mbox{12.64 ± 6.78}   & \textbf{\mbox{10.25 ± 3.63}} \\
               & \mbox{20~--~30}      & \mbox{11.01 ± 5.19}    & \mbox{8.17 ± 3.72}     & \mbox{8.06 ± 2.98}   & \textbf{\mbox{6.95 ± 2.24}} \\
               & \mbox{30~--~40}      & \mbox{8.89 ± 3.79}     & \mbox{7.49 ± 3.07}      & \mbox{7.42 ± 2.86}   & \textbf{\mbox{6.04 ± 2.09}} \\
\bottomrule
\end{tabularx}
% \caption{Your caption here}
\label{tab:visibility_tre}
\end{table}

% \begin{figure}[!ht]
%     \centering
%     \includegraphics[trim=0cm 0cm 0cm 0cm,clip,width=\linewidth]{pics/TRE_vs_visibility_nki.png}
%     \caption{Placeholder for visibility experiment.}
%     \label{fig:visibility}
% \end{figure}

% \begin{figure}[!ht]
%     \centering
%     \includegraphics[trim=0cm 0cm 0cm 0cm,clip,width=0.5\linewidth]{pics/TRE_vs_visibility.pdf}
%     \caption{Placeholder for visibility experiment.}
%     \label{fig:visibility_hhlbm}
% \end{figure}

Intraoperative liver surfaces are often only partially visible due to limited view angles and occlusion by instruments and adjacent organs, complicating the registration process. We carry out an experiment to examine the performance of the registration methods under various ranges of relative visibility, which is defined as $\alpha = \mathcal{A}(\mathbf{S}) / \mathcal{A}(\partial(\mathbf{V}))$, where $\mathcal{A}(\cdot)$ is a function to calculate the area of a given surface, $\partial(\cdot)$ extracts the surface mesh from a volume, and the visibility $\alpha \in [0,1]$. For this specific experiment we adopt the Phantom and HHLBM test sets. As described in Sec.~\ref{eval:datasets}, for each sample, both datasets contain $20$ partial intraoperative surfaces which are extracted from the view of a virtual camera, observing the anterior side of the liver volumes and gradually moving away from the liver.

For the Phantom dataset, at the most extreme occlusion ($0\text{–}10 \%$ visibility), PBSM yields the largest TRE ($22.30 \pm 6.83$\,mm), whereas our method achieves the lowest TRE ($9.33 \pm 4.23$\,mm), improving upon V2S-Net by $32\%$ and BCF\_FEM by $9\%$ of TRE. As the visibility increases, our method shows a declining trend and always obtains the best TREs around $6.5$\,mm with a notably lower variability, outperforming BCF\_FEM and V2S-Net by $2-3$\,mm. In comparison, the other methods present a fluctuating trend with the increase of visibility, stagnating around $9$\,mm TRE.

Regarding TREs on HHLBM dataset, under extreme occlusion ($0\text{–}10 \%$), V2S-Net and BCF\_FEM outperform our method ($18.09$\,mm and $18.61$\,mm vs. $19.93$\,mm for our method), suggesting that iterative optimization like BCF\_FEM can assist exploration of a better solution and voxelized data like the inputs of V2S-Net favor sensing the ambient small geometry. In contrast, our method performs only one forward end-to-end inference and introduces no extra spatial information. Once over $10\%$ visibility, the TRE of our method drops more quickly than the others and our method outperforms the baseline methods, with around $6$\,mm as the lowest error at the highest level of visibility. Overall, insights can be drawn from this experiment:

\textbf{Sensitivity to visibility}: All methods benefit from increased visibility ($\alpha$), but PIVOTS exhibits the steepest decline in TREs, indicating superior exploitation of additional surface data.

\textbf{Visibility Threshold}: $10\%$ visibility splits the performance of the methods on obtaining TRE lower than $1$\,cm. But above $10\%$, methods, especially PIVOTS, rapidly reduce TRE toward less than $7$\,mm. This indicates a critical visibility threshold that registration methods must aim to %exceed 
tackle for reliable accuracy in clinical setups.

% The results are listed in Tab.~\ref{tab:visibility_tre}. The X-axis shows the relative visibility and Y-axis shows the TRE. From the figure one can see both datasets contains partial surfaces with $\alpha \in (0.0, 0.45)$, proving the high difficulty of the two test datasets. 

% For both datasets, the TRE of P-V2S-Net decreases faster than the other methods with the increase of visibility, and the TRE of P-V2S-Net with largest visiblity amount range (0.3 - 0.45) outperforms all the other methods. With low visible surfaces, P-V2S-Net also achieved reasonably good results, proving the robustness against various visibility.

\begin{figure*}[!ht]
    \centering
    % \frame{
    \includegraphics[trim=0cm 13.8cm 1.5cm 0cm,clip,width=\textwidth]{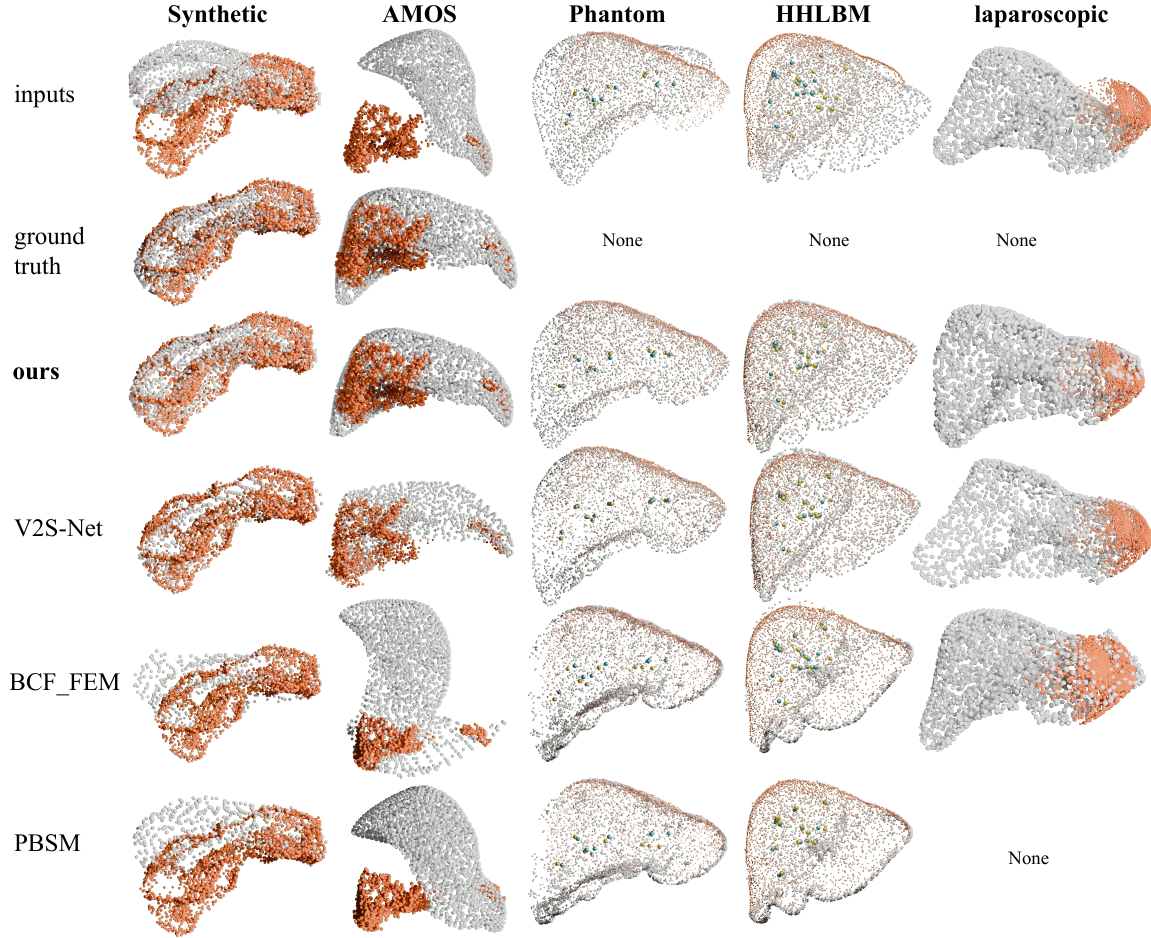}
    % }
    \caption{Qualitative comparison on various evaluation datasets. Preoperative point clouds are depicted in \colorbox{gray}{gray}, and intraoperative points in \colorbox{orange}{orange}. For Phantom and HHLBM datasets, preoperative landmarks are shown in \colorbox{yellow}{yellow} and intraoperative ones in \colorbox{cyan}{blue}. Ground truth information is only available for synthetic and AMOS datasets.}
    \label{fig:qualitative}
\end{figure*}

\subsection{Ablation Studies}
\label{sec:ablation}
\begin{table}[ht]
\centering
\caption{Ablation study on synthetic dataset}
\label{tab:ablation}
\begin{tabular}{lc}
\toprule
\textbf{Method} & \textbf{MED (mm)}  \\%  & \textbf{XXX} \\
\midrule
ours - 2 layers & 5.46  \\% & -- \\
ours - 4 layers & 4.89  \\% & -- \\
ours w/o deform att & 6.60  \\% & -- \\
ours w/o manual features & 4.50  \\% & -- \\
ours w/ KPConv encoder & 7.01  \\% & -- \\
ours & 3.90 \\% & -- \\

\bottomrule
\end{tabular}
\end{table}

In this section, we investigate various alternative architectures to %show the superior performance 
prove the necessity of the components of our proposed method. For the sake of saving training time, we train all models on a subset of the full generated training set, consisting of around one-third of the samples. 

\textbf{Number of cross-attention layers}: We show that increasing the number of layers can increase the registration performance, however, the gain decreases when more layers are involved, and the training time increases largely. Hence, the chosen number of six layers suits both our hardware setup at hand and registration performance.

\textbf{Deformation aware cross attention}: The counterpart of the proposed deformation aware cross attention is the interweaving attention, where the updated features are fed into cross attention between network layers. The $70 \%$ increase in registration error shows the pivotal influence of the deformation aware cross attention modules on the performance of the network, highlighting again the importance of designing the method specifically for this registration task.

\textbf{Manual feature extraction}: We show that without the manually extracted features, the network loses necessary information, so the performance degrades.

\textbf{KPConv encoder}: KPConv works as a successful point cloud encoder for various registration networks, e.g. LiverMatch. %Our earlier experiment adopts the design of kernel points from KPConv to ensamble our encoder. However, the inferior performance proves KPConv is not a suitable design for our task.
In this experiment, the design of kernel points is adopted from KPConv to enrich our encoder. However, despite helping rigid LiverMatch achieve results comparable to nonrigid methods in Tab.~\ref{tab:registration_performance}, the inferior performance in this case proves KPConv is not a suitable component for our architecture.

\subsection{Qualitative Results}
\label{exp:qualitative}

We visualize registration results of various methods on all test sets in Fig.~\ref{fig:qualitative}. The first row shows the input point clouds to the method, where the difference of PRDs between the datasets is clearly visible, and the second row contains ground truth for the synthetic and AMOS datasets.

\textbf{Synthetic data:} For the synthetic and AMOS datasets, BCF\_FEM and PBSM clearly struggle to accommodate large deformations: their output point clouds remain nearly in the initial pose, indicating a failure to drive the source toward the target geometry. On AMOS in particular, both methods leave the left lobe essentially uncorrected, and BCF\_FEM even produces very unrealistic deformation.
By contrast, V2S-Net matches the ground‐truth displacement well on the synthetic data, but on AMOS its deformation field exhibits abrupt changes. On the other hand, PIVOTS delivers much smoother and coherent displacement fields on both datasets and aligns the source point clouds well with the ground truth, providing an example for its superior quantitative performance.

\textbf{Real data:} For the Phantom and HHLBM samples, surface alignment improves for both BCF-FEM and PBSM relative to the synthetic test sets, aligning well with their gain in quantitative scores on real data in Tab.~\ref{tab:registration_performance}. Nonetheless, none of the two methods fully corrects the Phantom sample’s left‐lobe misalignment, and PBSM only slightly outperforms BCF-FEM on the right side. 
In the HHLBM example, PBSM yields the best result for the surface alignment and the landmarks, especially for the right lobe, and provides a more uniform misalignment vector distribution than V2S-Net. BCF\_FEM achieves a good local fit at the center but remains the least effective overall, showing no clear deformation trend. While both V2S-Net and PIVOTS succeed in moving internal landmarks closer to their true positions, PIVOTS exhibits stronger results in both the center / left of the liver and the surface regions.

\textbf{Intraoperative data:} The Laparoscopic Liver dataset allows only qualitative comparison but provides a valuable real intraoperative setting. 
% The deformation magnitude is similar to the synthetic example, surpassing the other two real datasets. 
BCF-FEM yields a plausible close-to-rigid transformation for the majority of the surface but it locally deforms the left lobe of the preoperative surface unrealistically. On the other hand, managing the alignment of the surfaces in the visible region well, only leaving a gap on the leftmost lower part, V2S-Net introduces a strong deformation to the right lobe with a sharp transition without a basis in the intraoperative information. PIVOTS predicts the most coherent deformation without local distortions and matches the partial surface very closely. It also assumes a deformation of the right lobe without a clear indication in the data, however, the displacement is in line with the transformation towards the target area. Since PBSM requires a surface mesh of the intraoperative data and generating such a mesh from the noisy point cloud introduces many additional parameters, highlighting a shortcoming of the method, we exclude it from this experiment.
% partial view of the intraoperative liver, containing noise from the stereo reconstruction as well as alignment errors due to deformation

The qualitative results highlight the ability of PIVOTS to predict consistent displacement fields across various amounts of deformation, while V2S-Net struggles with abrupt field changes and the biomechanical methods tend to introduce unrealistic deformations, especially at a higher discrepancy between source and target.

\section{Discussion and Conclusion}
% \(\rightarrow\) derive and sum up strengths and weaknesses of our method, identify open avenues and discuss why we can't fix those things in this paper yet\\
% \(\rightarrow\) how well can we answer the open question using this method? How does it compare to the other methods, everything taken together?\\
% - don’t forget to mention that input needs to be centered (easy enough)\\

% What we can't solve right now:\\
% - (n) different noise types from different input modalities \(\rightarrow\) In the future, a learnable noise-removal that preprocesses data from different inputs might be good.\\

% In this work, we propose a point cloud non-rigid liver registration network. This network consists of an encoder with DGCNN and FPS layers, and a decoder with \textit{deformation aware cross attention} and \textit{upsampling cross attention} modules. The proposed network achieves state-of-the-art performance on this difficult liver volume-to-surface registration task. 
\paragraph{Discussion}

Our comparison experiment shows that our proposed network not only outperforms biomechanics-based methods like PBSM and BCF\_FEM, but also existing deep learning-based methods, e.g. V2S-Net and Lepard, although these methods are re-trained on the same training sets with hyperparameter tuning. Moreover, our method generalizes well across datasets in terms of 1) shape variety: the model trained on synthetic data can seamlessly transfer to inference on AMOS samples, and 2) real datasets: PIVOTS maintains registration errors well below the 1\,cm threshold on unseen real-world Phantom and HHLBM datasets, proving its applicability in surgical scenarios.

The robustness of PIVOTS extends to large deformations (0\,mm–100\,mm PRD), where it achieves less than half the TRE of the next best method and three‐ to six‐fold improvements over other baselines. Its noise resilience is equally impressive: While V2S-Net’s error distribution broadens with increasing perturbation, PIVOTS preserves a tight error profile across both Gaussian and Perlin noise regimes. Furthermore, when faced with extremely limited fields of view (< $10\%$ visible surface) on HHLBM samples, our method initially lags behind baseline methods, however, performance rapidly improves with just $10\%$–$20\%$ visibility, achieving less than $1$\,cm errors across two real datasets. Additionally, PIVOTS exhibits low variability across patients and experimental conditions, ensuring reliable performance not just on average but for every individual case, which is also a critical requirement for clinical deployment.

We attribute this to the dedicated design of the network architecture (proved in the ablation study) and the training data, keeping the special characteristics of the practical surgical application in mind at all times.

Nevertheless, limitations still exist, hindering the translation into the operation room. Our method assumes the availability of a dense intraoperative point cloud. However, real-time 3D reconstruction in the operating room remains challenging. Recent advances in intraoperative SLAM \citep{Docea2021}, Neural Radiance Fields (NeRFs) for dynamic scenes \citep{Khojasteh2025}, and learning-based deformation tracking \citep{Gong2024} offer promising directions, but fully robust, low-latency reconstruction pipelines are not yet standard clinical tools. 

Like other learning-based 3D–3D non-rigid registration approaches (e.g., V2S-Net), our method presumes an initial rigid alignment of preoperative and intraoperative data. This requirement can limit its applicability in extreme scenarios where only a wrong or no initial pose estimate is available. Future work could integrate a rigid registration module to enable fully end-to-end alignment from arbitrary initial poses.

In our visibility experiments, all methods, including PIVOTS, struggle when less than $10\%$ of the surface is visible. This finding indicates that pipelines for intraoperative surface acquisition must prioritize wide coverage: not only during initial mapping but also through continuous map maintenance, reintegration of temporarily occluded regions, and real-time update of moving structures. Without sufficient coverage, even the most advanced registration networks cannot compensate for missing geometry.

The noticeable sim-to-real gap also calls for further improvements. This could be addressed by creating more realistic synthetic data modelling tool–tissue interaction, self-collision, heterogeneous material properties, volume changes due to pneumoperitoneum, and authentic surgical noise, and the effect of including each aspect would have to be evaluated. Training data efficiency could be increased by exploring adaptive sampling strategies. Gathering high-fidelity intraoperative datasets will be crucial for validating PIVOTS under true clinical conditions.

Finally, even a perfect non-rigid registration of preoperative information onto the intraoperative situation can only visualize what is discernible in the preoperative scan. In many cases, especially small lesions are first detected intraoperatively \citep{chu_current_2023, husarova_intraoperative_2023}. Therefore, the further advancement of reliable intraoperative imaging remains indispensable for improved patient outcome, while navigation using non-rigid registration algorithms like PIVOTS aids in contextualizing the resulting captures to transfer them into actionable next steps.

\paragraph{Conclusion}

In this work, we propose a point cloud non-rigid liver registration network. This network consists of an encoder with DGCNN and FPS layers, and a decoder with \textit{deformation aware cross attention} and \textit{upsampling cross attention} modules. Through extensive experiments, we demonstrate that the proposed network achieves state-of-the-art performance and outperforms baseline methods on the difficult liver volume-to-surface registration task. Moreover, it presents significant robustness against large deformation, substantial amounts of noise, and various intraoperative visibility, making it a strong candidate for non-rigid registration at the core of an intraoperative navigation system. Finally, in light of the generalizability of our method across random shapes and the easy adaptation of the training data generation to different tissue properties, we trust that the proposed method can be applied in deformation prediction of many other soft tissues, such as the prostate, lungs, etc. 

Future work will focus on improving the registration accuracy on low-visibility surfaces, which could be done by introducing a more powerful point cloud encoder for better geometry information extraction. Another challenging future task is further reducing the sim-to-real domain gap, which can be achieved via more realistic synthetic data generation with surgical scene understanding and dedicated data collection for validation. Additionally, more datasets of other soft tissues can be gathered to evaluate the generalizability of the proposed registration network.

% - in a nutshell, we present a specifically designed, powerful (= high accuracy) and robust method for volume-to-surface registration in surgery

% \paragraph{Outlook}
% - different encoding or relative coordinates to avoid centering the data

%%       \citet{<label>} ==> Lamport (1994)
%%       \citep{<label>} ==> (Lamport, 1994)
%%


\section*{CRediT authorship contribution statement}

\textbf{Peng Liu}: Writing – review \& editing, Writing – original draft, Visualization, Validation, Resources, Methodology, Investigation, Formal analysis, Data Curation, Conceptualization, Software.
\textbf{Bianca Güttner}: Writing – review \& editing, Writing – original draft, Conceptualization, Software, Validation, Formal analysis, Data Curation, Visualization.
\textbf{Yutong Su}: Visualization, Data curation, Methodology.
\textbf{Jinjing Xu}: Writing – review \& editing, Visualization.
\textbf{Chenyang Li}: Writing – review \& editing, Visualization.
\textbf{Mingyang Liu}: Validation, Methodology.
\textbf{Zhe Min}: Validation, Methodology.
\textbf{Andrey Zhylka}: Data curation, Resources.
\textbf{Jasper Smit}: Data curation, Resources.
\textbf{Karin Olthof}: Data curation, Resources.
\textbf{Matteo Fusaglia}: Data curation, Resources.
\textbf{Rudi Apolle}: Data curation, Resources.
\textbf{Matthias Miederer}: Data curation, Resources.
\textbf{Laura Frohneberger}: Data curation, Resources.
\textbf{Carina Riediger}: Conceptualization, Resources.
\textbf{Jürgen Weitz}: Conceptualization, Resources.
\textbf{Fiona Kolbinger}: Writing – review \& editing, Validation.
\textbf{Stefanie Speidel}: Writing – review \& editing, Validation, Resources, Investigation, Formal analysis, Project administration, Conceptualization, Supervision, Funding acquisition.
\textbf{Micha Pfeiffer}: Writing – review \& editing, Validation, Resources, Investigation, Formal analysis, Project administration, Methodology, Conceptualization, Supervision.

\section*{Declaration of competing interest}

The authors declare that they have no known competing financial interests or personal relationships that could have appeared to influence the work reported in this paper.

\section*{Acknowledgments}

This research is funded by the European Union through CloudSkin under grant agreement ID 101092646 and by the German Research Foundation (DFG, Deutsche Forschungsgemeinschaft) as part of Germany’s Excellence Strategy – EXC 2050/1 – Project ID 390696704 – Cluster of Excellence “Centre for Tactile Internet with Human-in-the-Loop” (CeTI) of Technische Universität Dresden.

\section*{Data availability}

The code and dataset can be accessed from \url{https://github.com/pengliu-nct/PIVOTS}.

\appendix
\section{Supplementary material}
% \section{Data Preprocessing for State-of-the-Art Method Comparison}
% Micha: Remind me: I need to write a paragraph about how I pre-processed the HHLBM and the NKI phantom data before we can run PBSM and BCF_FEM on them.

\bibliographystyle{elsarticle-harv} 
\bibliography{main.bib}



\end{document}

\section{Supplementary Material}

\subsection{Shape variability in generated training data}
\label{ssec:sup_shape_variability}
\begin{figure*}[!ht]
    \centering
    \includegraphics[width=1.0\textwidth]{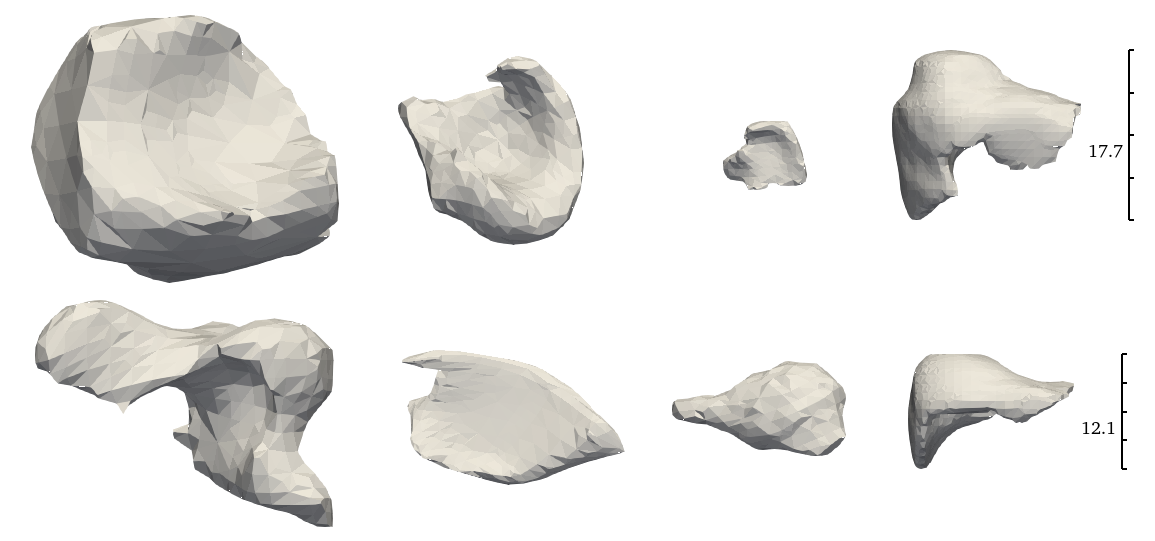}
    \caption{Shape variation in training data alongside two exemplary real livers.  Left: synthetic shapes, upper row: examples for different sizes, bottom row: examples for different shapes: concave, flat, and liver-like. Right: real livers (AMOS) and their variability. Bounding rulers show the liver size in superior-inferior direction in cm, all shapes shown in correct size relation.}
    \label{fig:sup_training_shapes}
\end{figure*}

\subsection{Data generation parameters}

%\centering
%\begin{table}[h!]
\begin{longtable}
    {%
        >{\raggedright\arraybackslash}p{0.3\linewidth}
        >{\raggedright\arraybackslash}p{0.4\linewidth}
        >{\raggedright\arraybackslash}p{0.3\linewidth}
    }
%\centering
\caption{Parameters used in the generation of the synthetic training data.}
\label{tab:sup_training_parameters} \\
%\begin{tabular}{>{\raggedright}p{0.3\linewidth} p{0.4\linewidth} p{0.3\linewidth}}
\toprule
\textbf{Pipeline block or scene object} & \textbf{Parameter} & \textbf{Value or range} \\
\midrule
\endfirsthead

\toprule
\textbf{Pipeline block or scene object} & \textbf{Parameter} & \textbf{Value or range} \\
\midrule
\endhead

\multicolumn{3}{r}{\textit{Continued on next page}} \\
\endfoot

\bottomrule
\endlastfoot

\multirow{9}{*}{Deformable organ} 
  & Young's modulus $E$        & $U(3, 30)$\,kPa \\
  & Poisson ratio $\eta$         & $U(0.45, 0.48)$ \\
  & mass density $\rho$          & $U(1050, 1090)\,\frac{\text{kg}}{\text{m}^3}$ \\
  & size          & $U(0.1, 0.3)$\,m \\
  & create concavities          & True \\
  & predeform twist          & True \\
  & predeform noise          & True \\
  & cut to fit          & True \\
  & voxel resolution          & $1$\,m \\
\midrule
\pagebreak

%\multicolumn{3}{@{}l}{
%\begin{minipage}{\linewidth}
%\begin{tabular}{@{}>{\raggedright\arraybackslash}p{0.3\linewidth} p{0.4\linewidth} p{0.3\linewidth}@{}}
\multirow{3}{*}{Organ placeholders} 
  & size          & $U(0.05, 0.2)$\,m \\
  & existence likelihood         & \parbox{\linewidth}{$0.9 \text{ for } i < 1$, $0.5 \text{ otherwise}$, $ 0 \leq i \leq 2$} \\
  & create concavities          & $P(\text{True})=0.5$ \\
%    \end{tabular}
%  \end{minipage}
%} \\

\midrule
\multirow{4}{*}{Ligaments} 
  & existence likelihood         & $1 \text{ for } i=0, ~~0.3 \text{ otherwise}$, $0 \leq i \leq 1$ \\
  & stiffness $k_{lig}$       & $U(100, 300)\,\frac{\text{N}}{\text{m}}$ \\
  & rest length factor $c_0$  & $U(0.9, 1.1)$ \\
  & path length       & $0.1$\,m \\
\midrule
\multirow{1}{*}{Fixed attachments} 
  & surface amount           & $U(2, 5)$\% \\
\midrule
\multirow{3}{*}{Abdominal wall} 
  & base outset amount           & $U(0, 0.05)$\,m \\
  & outset noise amplitude         & $U(0, 0.02)$\,m \\
  & outset noise frequency         & $U(1, 10)$ \\
\midrule
\multirow{2}{*}{Random scene block} 
  & extrusion size factor for deformable organ           & $0.5$ \\
  & target minimum voxel size (blender)         & $0.02$\,m \\
\midrule
\multirow{1}{*}{GMSH meshing block} 
  & maximum mesh element size           & $0.01$\,m \\
\midrule
\multirow{8}{*}{Simulation block} 
  & time step           & $50$\,ms \\
  & linear system solver         & Conjugate Gradient \\
  & maximum iterations         & $1000$ \\
  & minimum threshold         & $10^{-8}$ \\
  & tolerance         & $10^{-8}$ \\
  & time stepping scheme         & Implicit Euler \\
  & Rayleigh damping stiffness coefficient         & $0.1$ \\
  & Rayleigh damping mass coefficient         & $0.1$ \\
\midrule
\multirow{6}{*}{\parbox{\linewidth}{Random\\surface extraction block}} 
  & surface amount           & $U(10, 100)$\% \\
  & noise frequency         & $15$ in x, y and z direction \\
  & noise phase         & $U(0, 150)$ in x, y and z direction \\
  & geodesic distance weight         & $U(0.1, 1)$ \\
  & surface normal alignment weight         & $U(0.1, 1)$ \\
  & noise weight         & $U(0.1, 1)$ \\
\midrule
\multirow{4}{*}{\parbox{\linewidth}{Camera view\\surface extraction block}} 
  & mesh distance           & $U(0.05, 0.2)$\,m \\
  & window width         &  $U(0.03, 0.1)$\,m \\
  & aspect ratio         & $P(\text{16/9})=P(\text{4/3})=0.5$ \\
  & horizontal view angle         & $70\degree$ \\ %TODO degree sign?
\midrule
\pagebreak
\multirow{10}{*}{Surface noise block} 
  & Perlin noise amplitude $A_P$          & $U(0, 0.01)$\,m \\
  & Perlin noise frequency         & $U(10, 70)$ in x, y and z direction \\
  & Perlin noise phase         & $U(1, 999)$ \\
  & Gaussian noise $\sigma$         & $U(0, 0.003)$\,m \\
  & subdivision factor         & $U(5,15)$ \\
  & maximum subdivision triangle area         & $10^{-5}$ \\
  & maximum subdivision edge length        & $10^{-5}$ \\
  & sparsification Perlin noise frequency         & $U(1,5)$ \\
  & sparsification scale        & $1$ \\
  & sparsification shift        & $-0.3$ \\

%\end{tabular}
%\end{table}
\end{longtable}

\begin{longtable}
    {%
        >{\raggedright\arraybackslash}p{0.3\linewidth}
        >{\raggedright\arraybackslash}p{0.4\linewidth}
        >{\raggedright\arraybackslash}p{0.3\linewidth}
    }
%\begin{table}[h!]
%\centering
\caption{Parameters used in the generation of the AMOS validation and the AMOS$_{\text{noise}}$ data. Only parameters are listed that differ from the training data generation. Surface noise for the AMOS validation set was created as listed in table \ref{tab:sup_training_parameters} for the training data, for AMOS$_{\text{noise}}$ please refer to this table.}
\label{tab:sup_amos_parameters} \\
%\begin{tabular}
%    {%
%        >{\raggedright\arraybackslash}p{0.3\linewidth}
%        >{\raggedright\arraybackslash}p{0.4\linewidth}
%        >{\raggedright\arraybackslash}p{0.3\linewidth}
%    }
\toprule
\textbf{Pipeline block or scene object} & \textbf{Parameter} & \textbf{Value or range} \\
\midrule
\endfirsthead

\bottomrule
\endlastfoot

\multirow{1}{*}{Ligaments} 
  & rest length factor $c_0$  & $U(0.7, 1.1)$ \\
\midrule
\multirow{1}{*}{\parbox{\linewidth}{Scene object generator\\block}} 
  & source mesh scale factor           & $10^{-3}$ \\
  & & \\
\midrule
\multirow{4}{*}{Random scene block} 
  & blender remeshing octree depth        & $5$ \\
  & blender remeshing scale  & $0.75$ \\
  & mesh subdivision       & False \\
  & remove non-manifold regions   & True \\
\midrule
\multirow{2}{*}{Surface noise block} 
  & Perlin noise amplitude $A_P$          & $\in \{0, 3, 6, 9, 12, 15\}$\,mm \\
  & Gaussian noise $\sigma$         & $\in \{0, 2.5, 5\}$\,mm \\
%\bottomrule
%\end{tabular}
%\end{table}
\end{longtable}

\newpage
\subsection{Phantom test dataset collection}
\begin{figure}[!ht]
    \centering
      \includegraphics[width=0.7\linewidth]{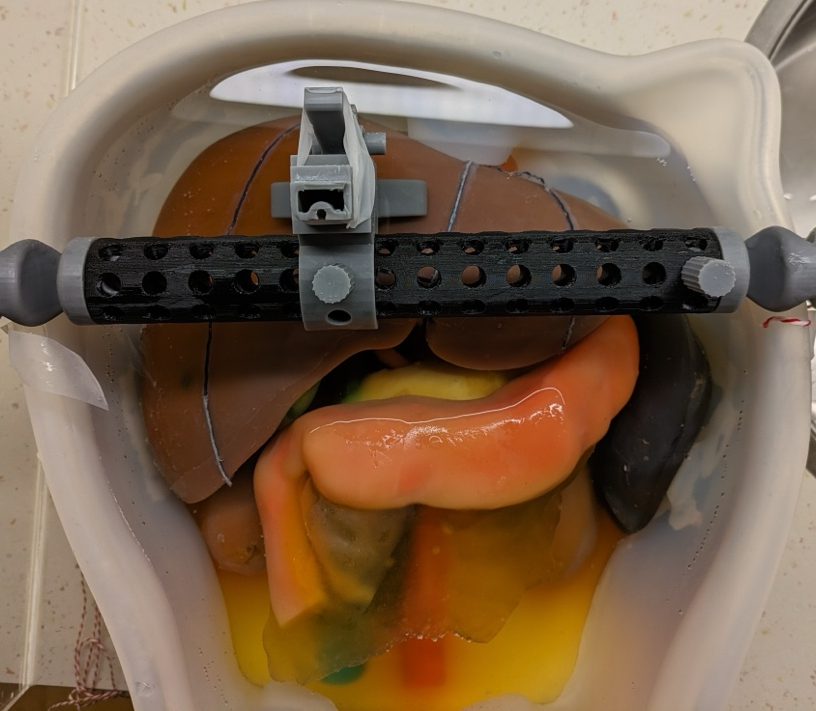}
    \caption{Phantom setup before MRI acquisition. The rail allows to fix the fake US probe and maintain the deformation during the scan.
    }
\label{fig:phantom}
\end{figure}